\documentclass[journal]{IEEEtran}
\usepackage{amsmath,amsfonts}
\usepackage{algorithmic}
\usepackage{algorithm}
\usepackage{array}
\usepackage[caption=false,font=normalsize,labelfont=sf,textfont=sf]{subfig}
\usepackage{textcomp}
\usepackage{stfloats}
\usepackage{url}
\usepackage{verbatim}
\usepackage{graphicx}
\usepackage{cite}

\usepackage{amsthm}
\usepackage{mathabx}
\usepackage{multirow}
\usepackage{float}

\newtheorem{theorem}{Theorem}

\newtheorem{assumption}{Assumption}
\usepackage{array} 
\usepackage{booktabs}
\usepackage{tabularx}
\usepackage{orcidlink} %
\hypersetup{hidelinks} % hide the hyperlink box      

\begin{document}

\title{EASTER: Embedding Aggregation-based Heterogeneous Models Training in Vertical Federated Learning}

\author{Shuo~Wang$^{\orcidlink{0009-0006-5388-2774}}$,
       Keke~Gai$^{\orcidlink{0000-0001-6784-0221}}$,~\IEEEmembership{Senior Member, IEEE}, 
       Jing~Yu$^{\orcidlink{0000-0002-3966-511X}}$,~\IEEEmembership{Member,~IEEE}, 
       Liehuang~Zhu$^{\orcidlink{0000-0003-3277-3887}}$,~\IEEEmembership{Senior~Member,~IEEE}, 
       Weizhi~Meng$^{\orcidlink{0000-0003-4384-5786}}$,
       ~\IEEEmembership{Senior~Member,~IEEE},
       Bin~Xiao$^{\orcidlink{0000-0003-4223-8220}}$,~\IEEEmembership{Fellow,~IEEE}

\IEEEcompsocitemizethanks{
\IEEEcompsocthanksitem S. Wang, K. Gai, and L. Zhu are with the School of Cyberspace Science and Technology, Beijing Institute of Technology, Beijing 100081, China, and K. Gai is also with the School of Artificial Intelligence, Beijing Institute of Technology, Beijing 100081, China. (e-mails: \{3220215214, gaikeke, liehuangz\}@bit.edu.cn). 
\IEEEcompsocthanksitem J. Yu is with the Key Laboratory of Ethnic Language Intelligent Analysis and Security Governance of MOE, Minzu University of China, and is also with the School of Information Engineering, Minzu University of China. (Email: jing.yu@muc.edu.cn).
\IEEEcompsocthanksitem W. Meng is with School of Computing and Communications, Lancaster University, United Kingdom. (Email: weizhi.meng@ieee.org).
\IEEEcompsocthanksitem B. Xiao is with the Department of Computing, The Hong Kong Polytechnic University, Hong Kong, China (e-mail: b.xiao@polyu.edu.hk).
\IEEEcompsocthanksitem Keke Gai (gaikeke@bit.edu.cn) is a co-first author; Jing Yu (jing.yu@muc.edu.cn) is the corresponding author.
}

\thanks{This work is supported by the National Key Research and Development Program of China (Grant No.s 2023YFF0905300), and the National Natural Science Foundation of China (Grant No.s U24B20146, 62372044), and Beijing Municipal Science and Technology Commission Project (Z241100009124008).}
}

\maketitle

\begin{abstract}
Vertical Federated Learning (VFL) allows collaborative machine learning without sharing local data. 
However, existing VFL methods face challenges when dealing with heterogeneous local models among participants, which affects optimization convergence and generalization of participants' local knowledge aggregation.
To address this challenge, this paper proposes a novel approach called \textit{\underline{E}mbedding \underline{A}ggregation-based Heterogeneou\underline{s} Models \underline{T}raining in Vertical F\underline{e}derated Lea\underline{r}ning} (EASTER).
EASTER focuses on aggregating the local embeddings of each participant's knowledge during forward propagation.
We propose an embedding protection method based on lightweight blinding factors, which injects the blinding factors into the local embedding of the passive party.
However, the passive party does not own the sample labels, so the local model's gradient cannot be calculated locally.
To overcome this limitation, we propose a new method in which the active party assists the passive party in computing its local heterogeneous model gradients.
Theoretical analysis and extensive experiments demonstrate that EASTER can simultaneously train multiple heterogeneous models and outperform some recent methods in model performance.
For example, compared with the state-of-the-art method, the model accuracy of EASTER was improved by 7.22\% under the CIFAR-10 dataset.
\end{abstract}


\section{Introduction} \label{sec: intro}
\IEEEPARstart{A}{s} the amount of data generated by mobile clients and Web-of-Things (WoT) devices increases, it drives the development of a wide range of machine-learning applications
Mobile clients and WoT devices have always had to share their data with the cloud to help with machine model training. 
Due to privacy concerns, mobile clients and WoT devices are wary about sharing raw data. 
As a new distributed machine learning paradigm, \textit{Federated Learning} (FL) \cite{pan2023fedmdfg} offers a solution for locally training ML models with data from mobile and WoT devices.
Particularly, FL implements a multi-participant collaborative training methodology that maintains data locally, ensuring data security.
Two major types of FL that have been paid wide attention in recent years include \textit{Horizontal Federated Learning} (HFL) \cite{so2023securing}, and \textit{Vertical Federated Learning} (VFL) \cite{gong2023multi}. 
Specifically, HFL enables collaborative training among participants sharing identical feature sets but having different sample spaces, whereas VFL addresses scenarios where participants hold distinct feature subsets of the same sample space \cite{aledhari2020federated}. 
This characteristic inherently leads to \textit{Non-Independent and Identically Distributed} (Non-IID) data distributions in VFL. 
Notably, the shared sample space makes VFL a primary choice for cross-organization collaboration, as opposed to HFL which requires overlapping features across participants \cite{li2021survey}. 
We focus on VFL in this work.

According to prior research \cite{xia2022cascade, wu2023falcon}, we categorize VFL participants into active and passive parties.
The active party owns the labels and features, while the passive party only owns the features.
\begin{figure}[t]
\centering
\includegraphics[width=0.5\textwidth]{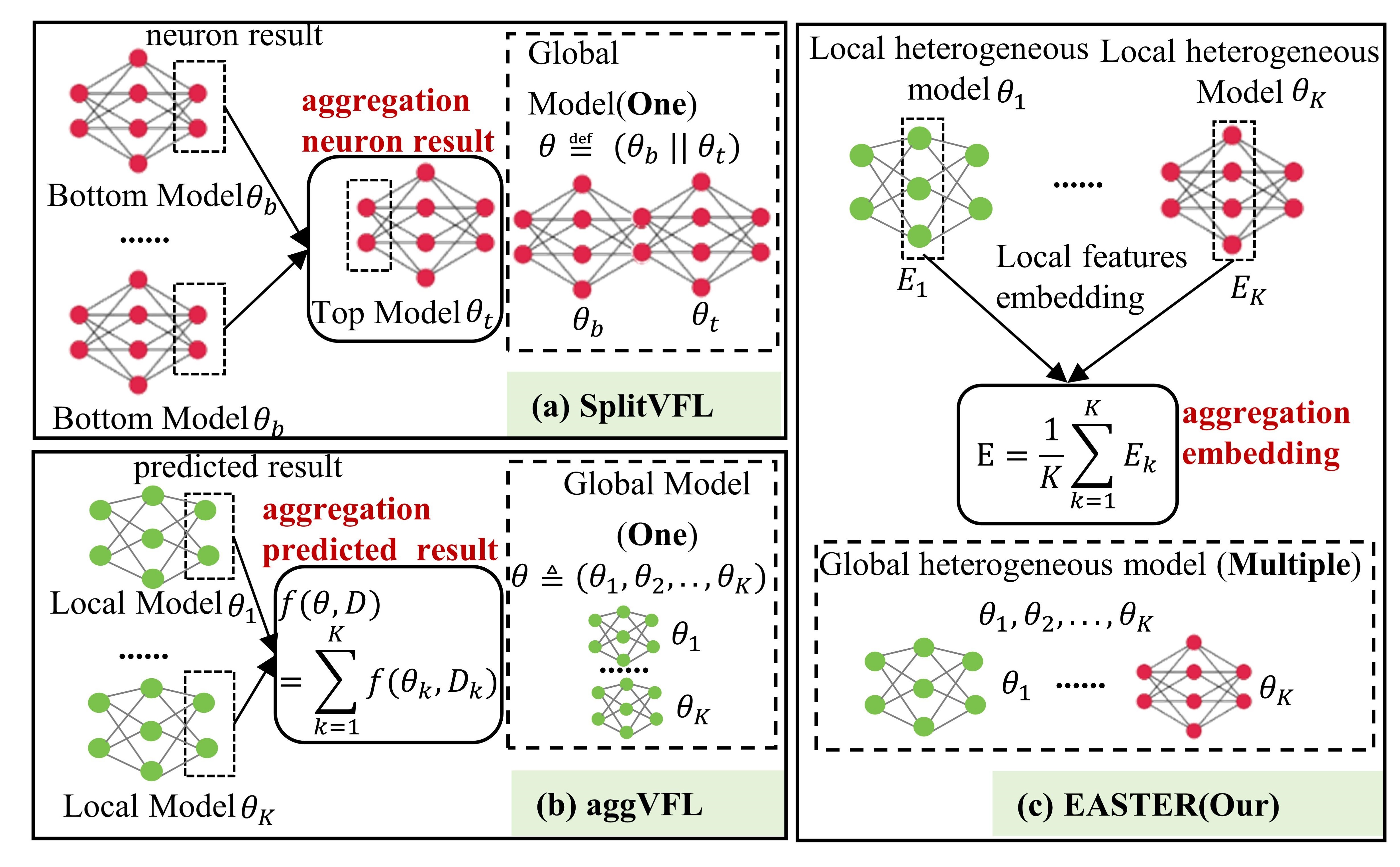} 
\caption{Typical architectures of VFL. (a): Current SplitVFL architectures; (b): Current aggVFL architectures; (c): The proposed VFL architecture (EASTER).
}
\label{fig: intro}
\end{figure}
Current research primarily divides VFL into SplitVFL \cite{jin2021cafe} and aggVFL \cite{wei2023fedads} based on the trainability of the top model.
The key difference between SplitVFL and aggVFL is that SplitVFL's aggregated top-level model is trainable, but aggVFL's aggregated model is not.
Specifically, SplitVFL as shown in Figure \ref{fig: intro}(a) splits a neural network model into a top and a bottom network.
The active party aggregates the outputs of the bottom-level networks to acquire all parties' local knowledge.
Meanwhile, the active party utilizes the aggregated neuron results for top-model training.
An aggVFL divides a network model into several parts, with each participant holding one part \cite{castiglia2022compressed}.
The active party aggregates the local prediction results of each participant.
It implies that most existing VFL research depends on an assumption that participating parties hold consistent local models with the same optimizer \cite{liu2022vertical}. 
As each participant has different resources and computing power, parties can choose varied local model architectures in most practical cases.
For example, some significant businesses or companies may have high-performance servers, GPU accelerators, or dedicated deep-learning accelerator cards, while others may have essential computer resources. 
Multiple medical institutions in the same area may have distinct medical image data that can be used to train a disease prediction model collaboratively. 
Hospitals in counties might only have essential computing resources, whereas hospitals in cities might have more advanced computing resources.
Therefore, in VFL, heterogeneity in computer resources is common. 
However, the heterogeneous model architecture of participants has been rarely addressed by prior studies \cite{liu2024vertical}. 
Specifically, the work \cite{liu2024vertical} provides a review of two VFL architectures, SplitVFL and aggVFL, and discusses their effectiveness, efficiency, and privacy considerations, but does not address optimization methods for heterogeneous models in privacy-preserving VFL.

To address model heterogeneity, the work \cite{ye2023heterogeneous} explores model-based approaches in HFL, including federated optimization, and cross-model knowledge transfer. These methods need participants can independently train local models. However in VFL, each participant only has a subset of features, limiting independent training of local models.
Our investigations find that existing work has rarely addressed heterogeneous VFL issues. 
Prior VFL works has evidenced that training the same model on varied datasets will output distinct performance in accuracy \cite{tan2022fedproto}.
Similarly, training different models on one dataset will receive varied accuracy performance \cite{fang2022robust, wang2023flexifed}. 
This phenomenon implies that model convergence is influenced by model heterogeneity and training datasets, though current methods mainly focus on scenarios with heterogeneous local models, neglecting heterogeneity and local knowledge from bottom model or forward propagation results.
Incorporating additional model information during VFL training may degrade the model's performance.

One of the keys to achieving VFL with heterogeneous local models is to reduce additional model information during aggregation.
However, methods for independent local models or gradient aggregation in HFL \cite{qi2024model} do not apply to VFL, as each participant in VFL only has a subset of features, making independent local model training impossible.
Inspired by embedding learning, heterogeneous neural networks can be divided into representation and decision layers \cite{qiao2023prototype}. 
The representation layer embeds similar features into the same embedding space, while the decision layer generates predictions from the embedded information and heterogeneous networks. 
Therefore, we observe that the embedding results obtained from the representation layer only contain the local knowledge of parties. 
Embedding learning provides a new perspective for achieving VFL with heterogeneous local models.

To tackle the model heterogeneous challenge in VFL, this paper proposes a method entitled \textit{\underline{E}mbedding \underline{A}ggregation-based Heterogeneou\underline{s} Models \underline{T}raining in Vertical F\underline{e}derated Lea\underline{r}ning} (EASTER).
Our method aggregates all parties' local embeddings to obtain global embeddings that contain the knowledge from the local training datasets of all parties.
These global embeddings are used for training heterogeneous local models by all participant parties. 
Considering the context of VFL, typically only active parties have access to labeled data and passive parties send intermediate forward propagation results to active parties. 
The active party assists passive parties in computing gradients to update heterogeneous local models, completing the local model training.

The main contributions of our work are summarized as 
\begin{itemize}
    \item We proposed a novel approach (e.g. EASTER) that aggregates local embeddings from multiple heterogeneous parties into a comprehensive global embedding.
    This method successfully optimizes various model architectures, achieving higher accuracy regardless of whether the local models are homogeneous or heterogeneous. 
    Moreover, EASTER is capable of simultaneously optimizing multiple heterogeneous models, resulting in several independently optimized models that maintain their distinct characteristics while benefiting from shared, aggregated insights. This capability significantly enhances the applicability and flexibility of VFL systems across diverse computational and data environments.
    \item 
    We proposed a secure embedding aggregation scheme within the EASTER framework that protects the privacy of local embeddings from participating parties. 
    This scheme rigorously protects local embedding privacy, ensuring the integrity and confidentiality of participant data during aggregation.
    As a result, our approach strengthens the security of the VFL system.
    \item
    We conducted a theoretical analysis and extensive experiments on six datasets to evaluate the EASTER performance. The evaluation metrics included model accuracy, communication overhead, computational and memory overhead, and the effect of heterogeneous participants on model performance.
    Compared with existing FL methods, EASTER has superior learning performance under heterogeneous local models. For example, compared with the state-of-the-art method, the model accuracy of EASTER was improved by 7.22\% under the CIFAR-10 dataset. 
\end{itemize}

The rest of this paper is organized in the following order.
Section \ref{Sec: Pre} presents the preliminaries of our method, essential for understanding the subsequent sections.
The system design and our proposed methodology are presented in Section \ref{sec: system} and Section \ref{sec: method}, respectively.
The experience evaluations in Sections \ref{sec: exper} and Section \ref{sec: dis} presents the discussion of our method.
In Section~\ref{sec: related}, we briefly review the related literature. 
Finally, we summarize this work in Section \ref{sec: coc}.

\section{Preliminaries} \label{Sec: Pre}
In this section, we provide some preliminary understanding of EASTER.
To simplify reading, we present descriptions for some notations in Table \ref{table: notation}.

\begin{table}[!t]
\centering
\caption{Notations Table.}
\begin{tabularx}{\linewidth}{l X}
\hline
\textbf{Notation}                 & \textbf{Explanations} \\
\hline
$\theta$         & Global model parameters \\
$\theta_b$       & Bottom model parameters  \\
$\theta_t$       & Top model parameters    \\
$\theta_k$       &  $l_k$th party's heterogeneous model parameters   \\
$E_{l_k}$           & Local embedding value generated by party $l_k$   \\
$[E_{l_k}]$        & Blinded local embedding value generated by party $l_k$   \\
$E$ & Global embedding\\
$l_k$ & The $k$th participant, and the special $l_0$ represents the active party\\
$N$ & The number of sample space $\mathbf{ID}$ \\
$K$ & the number of all passive parties \\
$C$ & the number of all parties \\
$\theta_{{l_k}}^{t}$ & Heterogeneous model parameters obtained by $l_k$th party in $t$th epoch \\
$f_{l_k} (\cdot)$ & The $l_k$th loss function \\
$D$ & Data sets participating in training    \\
$R_{l_k}$ & Local prediction value generated by passive party $l_k$\\
$\mathbb{G}$ & The cyclic group of prime order $p$\\
${SK}_{l_k}$& The privacy key generated by the $l_k$th passive party\\
${PK}_{l_k}$& The public key generated by the $l_k$th passive party  \\ 
$H(\cdot)$ & The collusion-resistant secure hash function \\
$r_{l_k}$ & Blinding factor \\
$\theta_k^*$ & The converged model parameters \\
\hline
\end{tabularx}
\label{table: notation}

\end{table}
\subsection{Vertical Federated Learning } \label{subsec: VFL}
VFL is a distributed machine-learning architecture that enables multiple clients to train a model collaboratively through sample alignment and feature union.
In VFL, a single party cannot train a complete model since labels and features are stored in multiple parties.
VFL aims to train a global model $\theta$ with multiple participants to minimize loss values without revealing local data.

The training process of VFL is as follows: (1) Each passive party uses local data features to perform local model forward training to obtain intermediate results, and uploads the intermediate results to the active party.
(2) The active party aggregates the intermediate results and calculates the global model loss.
(3) Each passive party downloads corresponding gradients from the active party to update their local model.

The training problem of VFL can be formulated as follows.
\begin{equation}\label{eq: VFLproblem}
    \min _{\theta \in \mathbb{R}^{d}} \ell(\theta, D): \triangleq \frac{1}{N} \sum_{j = 1}^{N} f \left(\theta ; \mathbf{x}_{j}, y_{j}\right) 
    %+ \lambda \sum_{ k = 1}^{K} z\left(\theta^{k}\right)
\end{equation}

In Eq. (\ref{eq: VFLproblem}), $N$ is the total number of data samples. $\mathbf{x}_{j}$ denote the union of the data features of $j$th row for all passive parties. $y_{j}$ is the $j$th labels of sample. 
$\theta \in \mathbb{R}^{d}$ denotes the global parameters. $f (\cdot)$ is the loss function.

\subsection{Diffie-Hellman Key Exchange}
In this paper, we use the \textit{Decisional Diffie-Hellman} (DDH) key exchange to generate blinding factors that further protect the passive party's local embedding value.
Let $\mathbb{G}$ be a cyclic group of prime order $p$, and $g$ is a generator of $\mathbb{G}$. 
We assume the DDH problem is hard \cite{ma2023flamingo}.
Passive parties can securely share a secret as follows:
The one passive party obtains a secret $a$ and sets its public value to $g^a \in \mathbb{G}$. 
The other passive party obtains his secret $b$ and sets his public value to $g^b \in  \mathbb{G}$. 
The one passive party and the other passive party exchange the public values and raise the other party’s value to their secret, i.e., $g^{ab} = (g^a)^b = (g^b)^a$. 
If DDH is hard, only the two passive parties know the shared secret.

\section{Model Design} \label{sec: system}

\subsection{Design Goals}
To address the heterogeneous model challenge in VFL, we aim to design a VFL with multiple heterogeneous models.
The main design goals of EASTER are as follows:
\begin{itemize}
    \item \textbf{Heterogeneous Parties Collaboration}. EASTER should be able to achieve collaborative training of participants with heterogeneous models.
    That is, EASTER can aggregate heterogeneous model information as little as possible during the training process to achieve collaborative training of participants with heterogeneous models.
    \item \textbf{Multiple Models Training}. EASTER should be able to support training multiple heterogeneous models simultaneously. In one training, we should be able to obtain multiple optimized heterogeneous models. If there are currently three participants with heterogeneous models participating in VFL training, then at the end of one collaborative training, EASTER should be able to obtain three optimized heterogeneous models.
    \item \textbf{Privacy Protection}. EASTER should safeguard passive parties' local privacy and security. Participants cannot infer the original features from the obtained embeddings.
\end{itemize}

\subsection{Problem Statement} \label{subsec: Pro}
Under the EASTER setting, we consider $K$ passive parties and one active party.
We define $C$ as the number of all parties, and $C = K + 1$.
This paper focuses on the training process of VFL so we assume that both the active party and $K$ passive parties have the same sample space $\mathbf{ID}$.
We give the training set $\left\{\mathbf{x_i}, y_{i}\right\}_{i = 1}^{N}$, where $N$ denotes the number of sample space $\mathbf{ID}$, $\mathbf{x_i}$ denotes all features of $i$th sample and $y_{i}$ denote the lable of $i$the sample.
For a VFL system, each $\mathbf{x_i}$ is vertically distributed $C$ shares and stored in each party, i.e., $\mathbf{x_i} = \{\{x_i\}_a, \{x_i\}_{{l}_{1}}, \{x_i\}_{{l}_{2}}, ...., \{x_i\}_{{l}_{K}}\}$.
In VFL, the active party owns the privacy dataset $D_a = \left\{ \{\{x_i\}_a\}_{i = 1}^{N}, Y = \{y_i\}_{i = 1}^{N}\right\}$, while each passive party possesses private data $D_{l_k}=\{\{x_i\}_{l_k}\}_{i = 1}^{N}$, where $l_k$ denotes the $k$ passive party. For convenience, we use $l_0$ to represent the active party $a$.

In VFL, the goal is to collaboratively train models $\theta$ among the active party and all passive parties, who have the same data $\mathbf{ID}$ and differential features. 
In addition, to protect data privacy, parties keep their data locally during training.
For VFL, only the active party possesses the label space, while passive parties only have some feature space ${\{x_i\}_{(l_k)}}_{i = 1}^{N}$ and do not have label space. 
Labels and features are stored in different parties in VFL.
A single passive party cannot train a complete model.
This paper assumes that each party can individually choose a locally trained model $\theta$. 
The goal of EASTER is to train multi-models ($\theta_a, \theta_{l_1}, ..., \theta_{l_K}$) jointly with all parties without revealing local samples, which aims to minimize the loss of all training samples.
The problem of EASTER can be formulated as

\begin{equation}\label{eq:1}
    \min _{\theta_a, ..., \theta_{l_K} \in \mathbb{R}^{d}} \ell(\theta_a,..., \theta_{l_K} ; D): \triangleq 
    \frac{1}{N} \sum_{i = 1}^{N} f \left(\theta_a, ..., \theta_{l_K} ; \mathbf{x}_{i}, y_{i}\right) 
\end{equation}

In Eq. (\ref{eq:1}), $N$ is the total number of samples, $\mathbf{x}_{i} = \bigcup_{k = 0}^{K} \{x_i\}_{l_k}$ denote $i$th sample features, $y_{i}$ is the label of $i$th sample labels.
$\theta_a \in \mathbb{R}^{d}$ denotes the model parameters owned by the active party and $\theta_{l_k} \in \mathbb{R}^{d}$ denotes the model parameters owned by $l_k$th passive party. $f (\cdot)$ is the loss function. 

To meet the real application requirements, this paper considers that the active party and the passive parties have independent local models, namely $(\theta_a, ..., \theta_{l_K})$. 
Each local model can be expressed as Eq. (\ref{eq:1}), that is, it has $C$ loss functions $(f_a(\cdot), ..., f_{l_K}(\cdot))$.

We focus on solving the issue of the active and passive parties working together to train the $C$ heterogeneous local model safely. 
That is, the active party and all passive parties collaborate to minimize all loss functions $(f_a(\cdot), ..., f_{l_K}(\cdot))$.

\subsection{Threat Model}

We assume the honest-but-curious threat model, which is widely used to investigate the privacy vulnerabilities of VFL systems \cite{ye2024vertical,li2024approaching,zhang2021secure,citation-key}.
The passive party is capable of accurately performing local pre-training and embedding calculations. 
However, it may attempt to infer the label information of the active party from the training process.
On the other hand, while the active party can correctly execute the global model training process, it may attempt to infer the private information of the passive party from their embeddings.
External attackers and malicious message tampering are also ignored in our method.
A secure communication channel is established between the active and passive parties, preventing external attackers from accessing or detecting the communication model details.

\section{Proposed Methodology} \label{sec: method}
\begin{figure*}[t]
\centering
\includegraphics[width=0.8\linewidth]{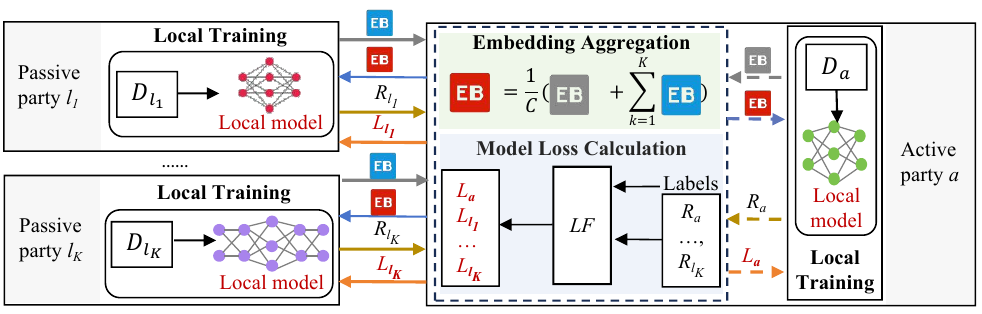} 
\caption{An overview of the EASTER training in the heterogeneous models setting. Each participant is the owner of the heterogeneous local model and incomplete features. The active party also has labels, partial features, and heterogeneous local models. $K$ passive parties and one active party are the suppositions made by EASTER. The dashed box is merely being used as an example. The solid box depicts the entity or module. 
}
\label{fig: EASTER}
\end{figure*}

\subsection{Overview} \label{subsec: promethod}
To solve the low accuracy problem caused by the passive party's local heterogeneous model, we proposed a new method EASTER.
An overview of the proposed EASTER is shown in Fig. \ref{fig: EASTER}.
Our scheme comprises $C$ participants, consisting of one active party $a$ and $K$ passive parties. 
Each participant owns private datasets and heterogeneous models locally. 
The main goal of EASTER is to aggregate the private datasets from all parties and utilize them collaboratively to train $C$ heterogeneous models.

The EASTER includes two entities: the active and passive parties.
In the training phase, an active party is mainly responsible for local heterogeneous model training, local embedding aggregation, and loss value calculation. 
Specifically, the active party collaborates with the passive party to train local heterogeneous models. 
The active party needs to aggregate the local embedding values of all passive parties and obtain the global embedding value to train heterogeneous models. 
The passive party is mainly responsible for initializing blinding factors and training local heterogeneous models. Specifically, the passive party generates a pair of public and private keys and sends the public key to the active party. The passive party generates a blinding factor based on its private key and the public keys of other passive parties to protect the local embedding value. 
In addition, the passive party completes the training of local heterogeneous models with the assistance of the active party.

In addition, the EASTER consists of three main modules. 
\begin{itemize}
    \item \textbf{Local Training}. All participants' collaborative training of multiple local heterogeneous models occurs in this module. 
    In detail, each participating party utilizes local features and models to generate local embedding values. Furthermore, each participant employs the global embeddings alongside their local models to derive local predictive outcomes Finally, each participant updates their local model in reverse according to the loss values and the optimization function.
    \item \textbf{Embedding Aggregation}. This module combines all participants' local embeddings to generate the global embedding.
    \item \textbf{Model Loss Calculation}. In this module, the active party aims to utilize local labels to support the passive party in calculating the heterogeneous model's loss value.
\end{itemize}

\subsection{Local Model Training}
To address the issue of a limited exchange of model information among participants, we adopted a strategy inspired by prior studies \cite{tan2022fedproto, tan2022federated}. 
When performing local training, the local heterogeneous network is partitioned into two distinct parts: the embedding network (e.g., embedding layer) and the prediction network (e.g., decision layers).
\textbf{The embedding layer} embeds some features owned by participants in the same embedding space. The embedding layer of the $l_k$th participant is $h(\theta_{l_k})$. We denote $E_{l_k} = h(\theta_{l_k}, D_{l_k})$ as the embedding of $D_{l_k}$ for ${l_k}$th participant.
\textbf{The decision layers} are the result of predicting sample $x$ based on the supervised learning task. We represent the prediction result by using $R_{l_k} = p(\theta_{l_k}, x)$.

\begin{figure}[t]
\centering
\includegraphics[width=1\linewidth]{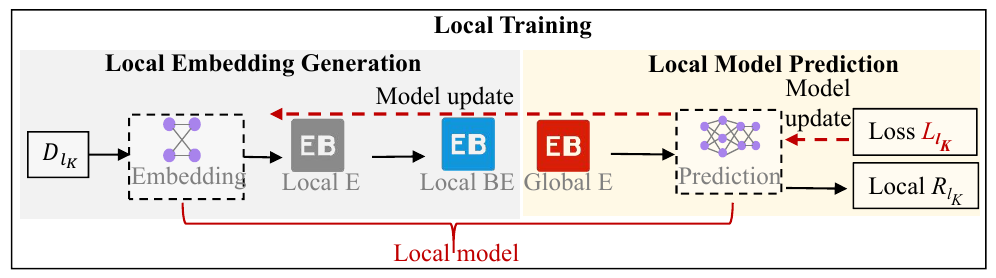} 
\caption{The local training process of EASTER.``Local E'' denotes the local embedding of the party. ``Local BE'' denotes the blinding local embedding of the passive party.
}
\label{fig: Localtraining}
\end{figure}

Local model training consists of local embedding generation, local model prediction, and model update, as shown in Fig.\ref{fig: Localtraining}.
Local embedding generation refers to the method by which participants acquire the local embedding result $E$ from the input layer to the embedding layer using the local heterogeneous network.
To project the local embedding result, we inject the blinding factors into the local embedding of the passive party.
Local model prediction refers to the process where each participant uses the global embedding values as inputs to the decision layer to obtain sample prediction values.
Model update is the process of optimizing model parameters from the output layer to the input layer using an optimization algorithm.
For example, when the $l_k$th participant gets the stochastic gradient descent (SGD) \cite{gao2021convergence} algorithms during backpropagation, the local heterogeneous model will be updated based on the following:
\begin{equation}
    \theta_{{l_k}}^{t+1} \leftarrow \theta_{{l_k}}^{t}-\eta_{{l_k}} \nabla_{\theta_{{l_k}}^{t}} L_{{l_k}}
\end{equation}
where $\theta_{{l_k}}^{t+1}$ stands for the model of ${l_k}$th participant in epoch $t$. $\eta_{{l_k}}$ denotes learning rate and $\nabla_{\theta_{{l_k}}^{t}} L_{{l_k}}$ denotes the gradient.
Below we introduce the process of local embedding blinding in detail.

\textbf{Local Embedding Blinding.}
In this paper, the global embedding aggregation is implemented in the active party, it implies that we only need to protect the local feature of the passive party.
We prevent the direct leakage of local features from computing the local embedding $\{{E_{l_k}}\}_{k = 1} ^ {K}$ at each participant locally.
To safeguard the original features further, we add a blinding factor ($r_{l_k}$) \cite{zheng2022aggregation} to the local embedding ($\{{E_{l_k}}\}_{k = 1} ^ {K}$) of the passive party to strengthen the security of aggregation. 
Each passive party initializes a public-private key pair and then generates the blinding factor.
\begin{itemize}
    \item \textbf{Key Generation.}
    Let $\mathbb{G}$ be a cyclic group of prime order $p$ with generator $g$, where the \textit{Discrete Logarithm Problem} (DLP) in $\mathbb{G}$ is computationally hard.
    Each $l_k$ generates a private key ${SK}_{l_k}=s_{l_k} \in \mathbb{Z}_p$ and a public key ${PK}_{l_k}=g^{s_{l_k}} \in \mathbb{G}$, where $k \in [1,K]$. 
    Each passive party sends its public key to the active party.
    \item  \textbf{Blinding Factor Generation.}
After the key initialization, each passive party $l_k$ downloads other passive parties' public key from the active party and computes the shared key $CK_{k,j}=H((PK_{l_j})^{SK_{l_k}})$, where $k,j \in [1, K],j \neq k$ and $H(\cdot)$ is a collusion-resistant secure hash function capable of converting strings of any length into elements in $\mathbb{Z}_p$.
The mathematical expression of the connection between shared keys is given in Eq. (\ref{eq:conn}).
\begin{equation}\label{eq:conn}
\begin{split}
CK_{k,j} & =  H({(PK_{l_j})}^{SK_{l_k}}) =H({(g^{s_{l_j}})^{s_{l_k}}}) \\
& = H({(g^{s_{l_k}})^{s_{l_j}}}) = H({(PK_{l_k})}^{SK_{l_j}}) = CK_{j,k}
\end{split}
\end{equation}
Each passive party $l_k$ computes the blinding factor to generate $r_{l_k}$ (see Eq. (\ref{eq: BF})). 
\begin{equation}
\label{eq: BF}
    r_{l_k}=\sum_{j \in [1,k], j \neq k}(-1)^{k>j} H({CK}_{k,j})
\end{equation} 
Where $(-1)^{k>j}=-1$ if $k>j$. % and 1 otherwise. 
Particularly, the sum of all blinding factors ($\sum_{k \in [1, K]} r_{l_k}$) is 0.
\end{itemize}

The local embedding value with privacy protection can be obtained by adding the above-mentioned blinding factor to the passive party's local embedding as follows.
\begin{equation}\label{eq:lse}
    \{[{E_{l_k}}]\}_{k = 1} ^ {K} = \{{E_{l_k}} + {r_{l_k}}\}_{k = 1} ^ {K}
\end{equation} 

The $l_k$th passive party sends the blinded embedding $[{E_{l_k}}]$ to the active party.
\subsection{Secure Embedding Aggregation} \label{subsubsec: embedding}
Given the heterogeneous model in participants, the optimal model parameters required by each participant are different. 
Existing aggregation-based VFL cannot provide enough effective information for each participant. 
Fortunately, with the same label features, the same embedding space can be obtained through the embedding layer of the heterogeneous network.
We effectively exchange the local features owned by aggregating the local embedding of each participant. 
The active party obtains the blinding local embedding, refer to $\{[{E_{l_k}}]\}_{k = 1} ^ {K} = \{{E_{l_k}} + {r_{l_k}}\}_{k = 1} ^ {K} $
A global embedding $E$ is generated after the average aggregating method (see Eq. (\ref{eq: E})).
\begin{equation} \label{eq: E}
    E = \frac{1}{C}  ( E_a + \sum_{k = 1}^{K}[E_{l_k}] )
     = \frac{1}{C}  (E_a + \sum_{k = 1}^{K}E_{l_k}+ \sum_{k = 1}^{K}r_{l_k}).
\end{equation}
In Eq. (\ref{eq: E}), $C$ denotes the total number of parties; $K$ denotes the total number of passive parties.
We know from blinding factor generation that $\sum_{k = 1}^{K}r_{l_k} = 0$, so we can obtain the global embedding.

\subsection{Model Loss Calculation}
In EASTER, the passive party lacks labels locally, making it unable to calculate the loss value independently. 
Only the active party possesses the labels and has a heterogeneous model loss value calculation module to aid the passive party in computing the loss value. 
The active party determines the optimal loss function $LF$ following the supervised learning task's requirements. 
When the active party selects the cross-entropy loss function, for instance, the loss function of the $l_k$th participant's heterogeneous model is calculated by Eq. (\ref{eq:lf}). 
\begin{equation} \label{eq:lf}
\resizebox{1.0\hsize}{!}{$
    LF\left(R_{l_k}, Y\right)=-\frac{1}{N} \sum_{i = 1}^{i = N}\left(\left(Y_{i}\right) \log_2 \left((R_{l_k})_{i}\right)+\left(1-Y_{i}\right) \log_2 \left(1-(R_{l_k})_{i}\right)\right)
$}
\end{equation}
In Eq. (\ref{eq:lf}), $R_{l_k}$ represents the predicted probability; $Y$ represents the actual label value; $N$ represents the total number of the training sample.

\begin{algorithm}[!t]
\caption{EASTER Training}
\label{alg: EASTER}
\begin{algorithmic}
\REQUIRE {Datasets $D_{l_k}$, parties $k = 0, 1, ..., K$, Epoch $T$}
\ENSURE {Excellent $\theta_{l_0}, \theta_{l_1}, ..., \theta_{l_K}$}
\end{algorithmic}
\begin{algorithmic}[1]
\FOR{$t = 1,...,T$}
\FOR{each party $l_k$ in parallel}
\STATE $[E_k]$ $\leftarrow$ $h(\theta_{l_k}, D_{l_k}) + r_{l_k}$  
\ENDFOR
\STATE Perform the global embedding $E$ update by Eq. (\ref{eq: E}) 
\STATE The active party sent $E$ to passive parties
\FOR{each party $l_k$ in parallel}
\STATE Perform local prediction $R_{l_k}$ = $p(\theta_{l_k}, E)$
\STATE Sent prediction $R_{l_k}$ to active party
\ENDFOR
\STATE $L_{l_k}$ = $LF$($R_{l_k}$, $Y$) // Calculate the loss of each party
\STATE The active party sent $L_{l_k}$ to $l_k$th passive party  
\FOR{each party $l_k$ in parallel}
\STATE $\theta_{l_k}^{t+1} \leftarrow \theta_{l_k}^{t}-\eta_{l_k} \nabla_{l_k} L_{l_k}$ // Party update local heterogeneous model
\ENDFOR
\ENDFOR
\end{algorithmic}
\end{algorithm}

\subsection{Multiple Heterogeneous Models Training}
Algorithm \ref{alg: EASTER} presents the pseudo-codes of EASTER for implementing multiple heterogeneous models training process. 
The party refers to the general terms of active party and passive parties.
$l_0$ represents the active party.
Participants only have the features subsets and need the assistance of other participants during the training process.
In addition, the local model of the passive party does not have labels and cannot calculate the loss value of the heterogeneous local model, which requires the cooperation of the active party. 

The training process of one round of EASTER mainly includes the following steps.
\textit{Step 1 } (line 3 to line 5 of the Algorithm \ref{alg: EASTER}) Each participant uses local data feature $D_{l_k}$ and local heterogeneous embedding network $h(\theta_{l_k}, D_{l_k})$ to obtain the local blinded embedding $[E_{l_k}] = h(\theta_{l_k}, D_{l_k}) + r_{l_k}.$ 
\textit{Step 2} (line 6) The active party realizes the safe aggregation of global embedding by Eq. (\ref{eq: E}) and obtains the local data characteristics of all participants.
\textit{Step 3} (lines 8 to 9) The participants calculate the prediction result $R_{l_k}$ from the global embedding E and the decision layer $p(\theta_{l_k}, E)$ of the heterogeneous network in parallel.
\textit{Step 4} (lines 11 to 13) The active party calculates the loss value of participant $l_k$ from the loss function $LF(R_{l_k}, Y)$, the prediction $R_{l_k}$, and local labels. 
$LF(R_{l_k}, Y)$ chooses different loss functions by the task requirements.
For example, the cross-entropy or the logistic regression loss function can be chosen for classification tasks. 
\textit{Step 5} (lines 13 to 15) Each participant updates and optimizes the parameters of the local heterogeneous model according to the loss and the optimization method (e.g., SGD, SGD with momentum, Adagrad, and Adam).
After $T$ training rounds, we will obtain $C$ local heterogeneous models.
Therefore, the EASTER model trains multi-heterogeneous models and safeguards the passive side's local embedding. EASTER enables the generation of multiple optimized heterogeneous models from a single training.

\subsection{Convergence Analysis} \label{subsec: analysis}
We provide the convergence analysis of EASTER below, which relies on the fact that the local gradients in the training process are unbiased. In our analysis, we use $f_k(\theta)$ to denote the loss function of $k$th participant. Additionally, we present a set of assumptions required to perform the convergence analysis, similar to existing general frameworks \cite{tan2022fedproto, zhang2021secure}.

\begin{assumption}[$L$ - Lipschitz Continuous] \label{assumption: LC}
    Suppose there exists a constant $L > 0$ and the gradient of the local objective function is $L$ - Lipschitz Continuous for $\forall$ $\theta_k, \theta_k'$ and $k \in \left\{0, 1,...,K \right\}$, there is
\end{assumption}
\begin{equation}
    \left \| \bigtriangledown f_k(\theta_k) - \bigtriangledown f_k(\theta_k')\right \|\le L\left \| \theta_k - \theta_k'\right \|
\end{equation}

\begin{assumption}[Unbiased Gradient and Bounded Variance] \label{assumption: UGBV}
    The local gradient $g_k^t = \bigtriangledown f_k(\theta^t, E)$ for each participant is an unbiased estimator, where $k$ denotes $l_k$th participant. 
    We assume that the expectation of the local heterogeneous model gradient $ g_k^t $ satisfies $\mathbb{E}\left[g_k^t\right]=\nabla f_k\left(\theta_k^t\right)=\nabla f_k^t, \forall k \in\{0,1,2, \ldots, K\}$. And the variance of the local gradient $ g_k^t$ is bounded by $\sigma_k^{2}$: $\mathbb{E}\left[\left\| g_k^t-\nabla f_k\left(\theta_k^t\right)\right\|\right] \leq \sigma_k^{2}, \forall k \in\{0,1,2, \ldots, K\}, \sigma_k^{2} \geq 0$.    
\end{assumption}

\begin{assumption}[Bounded Gradient] \label{assumption: BG}
    We assume that $G$ bounds the expectation of the local heterogeneous model gradient. We have
    $\mathbb{E}\left[\left\| g_k^t \right\|\right] \leq G, \forall k \in\{0,1,2, \ldots, K\}$.
\end{assumption}

\begin{assumption}[$u$ - convex] \label{assumption: LS}
    Suppose the $f_k$ is $u$ - convex, which means that there exists $u$ such that $\nabla^2 f_k(\theta) \geq u$
\end{assumption}
We aim to produce convergent outcomes for each participant's local heterogeneous model. 
When $\min_{\theta}$ is the smallest, we anticipate that each participant's objective function $f(\theta, D)$ converges. 
Since the proof of convergence for each participant's objective function is consistent, the convergence result of the statement for $k$th participant is displayed below. 
\begin{theorem}[Convergence of EASTER] \label{theorem: convergence}
    Under assumptions \ref{assumption: LC} - \ref{assumption: LS}, we prove that the convergence of EASTER is:
    \begin{equation}
    \begin{split}
    &\mathbb{E}\left[f_k\left(\theta_k^{t+1}\right)-f_k\left(\theta_k^{*}\right)\right] \leq \\
    &\left(1-\mu \eta_k \sigma_k^{2}\right) \mathbb{E}\left[f_k\left(\theta_k^{t}\right)-f_k\left(\theta_k^{*}\right)\right] 
        +\frac{1}{2} \eta_k^{2} L G
    \end{split}
\end{equation}
\end{theorem}

Where the $\theta_k^t$ represents the $k$th participant's local model obtained in the $t$ epoch and $\theta_k^*$ denotes the convergent model. $\eta_k$ denotes the learning rates in $t$th epoch. As the number of epoch $t$ increases, the upper bound of the distance between the current model parameter $\theta_k^t$ and the convergent model $\theta_k^*$ gradually decreases, which shows the convergence of the EASTER model.

Based on the existing assumptions, we provide complete proof of the convergence of EASTER.

\begin{proof}
We assume that Assumption \ref{assumption: LC} - \ref{assumption: LS} hold.
Since the local model in our method, EASTER, employs the SGD optimization algorithm, we draw upon the work of \cite{gao2021convergence}, which demonstrates the convergence properties of local SGD. Based on the four assumptions previously outlined, we derive the convergence results for the proposed EASTER method as follows:

At step $t$, we suppose the $k$th participant optimizes the model $\theta_k^t$. So the current step's stochastic gradient is $g_k^t$ (denotes $g$)and learning rate $\eta_k$. We have the following formula:
\begin{equation}\label{eq:9}
    \theta_k^{t+1}=\theta_{k}^t-\eta_{k} g\left(\theta_{k}^t\right)
\end{equation}

According to the Assumption \ref{assumption: LC}, the following inferences \ref{infere} will hold.
\begin{equation} \label{infere}
    f_k(\theta)-\left[f_k\left( \acute{\theta}\right)+\nabla f_k\left(\acute{\theta}\right)^{T}\left(\theta-\acute{\theta}\right)\right] \leq \frac{1}{2} L\left\|\theta-\acute{\theta}\right\|_{2}^{2}
\end{equation}

Eq. (\ref{eq:9}) can be expressed as follows based on Assumption \ref{assumption: LC} and Inferences \ref{infere}.

\begin{equation} \label{eq:11}
    f_k\left(\theta_{k}^{t+1}\right)-f_k\left(\theta_{k}^{t}\right)=-\eta_k \nabla f_k\left(\theta_k^{t}\right)^T g\left(\theta_k^{t}\right)+\frac{1}{2} L\left\|g\left(\theta_k^{t}\right)\right\|_{2}^{2}
\end{equation}
By taking the expectation of Eq. (\ref{eq:11}), the following equation is obtained:
\begin{equation}\label{eq:12}
\begin{split}
    \mathbb{E}\left[f_k\left(\theta_k^{t+1}\right)\right]-f_k\left(\theta_{k}^t\right) &\leq-\eta_k \nabla f_k\left(\theta_{k}^t\right)^{T} \mathbb{E}\left[g\left(\theta_{k}^t\right)\right] \\
&+\frac{1}{2} \eta_k^{2} L \mathbb{E}\left[\left\|g\left(\theta_{k}^t\right)\right\|_{2}^{2}\right]
\end{split}
\end{equation}

Based on Assumption \ref{assumption: UGBV} and Assumption \ref{assumption: BG}, Eq. (\ref{eq:12}) can be represented by the following equation.
\begin{equation} \label{eq:13}
\begin{split}
     A=-\eta_k \nabla f_k\left(\theta_{k}^t\right)^{T} \mathbb{E}\left[g\left(\theta_{k}^t\right)\right] \leq-\eta_k \sigma_k^{2}\left\|\nabla f_k\left(\theta_k^{t}\right)\right\|_{2}^{2}  \\
B=\frac{1}{2} \eta_k^{2} L \mathbb{E}\left[\left\|g\left(\theta_{k}^t\right)\right\|_{2}^{2}\right] \leq \frac{1}{2} \eta_k^{2} L(G \| \nabla f_k\left(\theta_k^{t}\right) \|^{2}+G)
\end{split}   
\end{equation}

According to Eqs. (\ref{eq:13}) - (\ref{eq:12}), an equivalent equation is obtained.

\begin{equation}
\resizebox{1.0\hsize}{!}{$
    \mathbb{E}\left(f_k\left(\theta_k^{t+1}\right)\right)-f_k\left(\theta_k^{t}\right)\leq 
    -\left(\eta_k \sigma_k^{2}-\frac{1}{2} \eta_k^{2} L G\right)\left\|\nabla f_k\left(\theta_k^{t}\right)\right\|^{2}
    +\frac{1}{2} \eta_k^{2} L G
$}
\end{equation}
The objective function should decrease at each step, which requires ensuring that the coefficient of $\|\nabla f_k\left(\theta_k^{t}\right)\|_{2}^{2}$ is less than 0. 
Therefore we need to assume $(\eta_k \sigma_k^{2}-\frac{1}{2} \eta_k^{2} L G)> 0$. 
For convenience, we assume $\sigma_k^2\leq \eta_k LG$. 
Upon repeated iteration, the ensuing equation shall be derived.
\begin{equation}
\resizebox{1.0\hsize}{!}{$
\mathbb{E}\left[f_k\left(\theta_k^{t+1}\right)\right]-f_k\left(\theta_k^{t}\right) \leq-\frac{1}{2} \eta_k \sigma_k^{2}\left\|\nabla f_k\left(\theta_k^{t}\right)\right\|_{2}^{2}+\frac{1}{2} \eta_k^{2} L G
$}
\end{equation}
The following formula could be obtained based on Assumption \ref{assumption: LS}.
\begin{equation} \label{eq:16}
\mathbb{E}\left[f_k\left(\theta_k^{t+1}\right)\right]-f_k\left(\theta_k^{t}\right) \leq
    -\mu \eta_k \sigma_k^{2}\left(f_k\left(\theta_k^{t}\right)-f\left(\theta^{*}\right)\right) \\
    + \frac{1}{2} \eta_k^{2} L G
\end{equation}
The goal of our optimization is to minimize $f_k\left(\theta_k^{t}\right)-f\left(\theta^{*}\right)$, and the Eq. (\ref{eq:16}) can be derived iteratively as follows.
\begin{equation} \label{eq:17}
    \begin{split}
       & \mathbb{E}\left[f_k\left(\theta_k^{t+1}\right)-f_k\left(\theta_k^{*}\right)\right]-\mathbb{E}\left[f_k\left(\theta_k^{t}\right)-f_k\left(\theta_k^{*}\right)\right] \leq\\
        &-\mu \eta_k \sigma_k^{2} \mathbb{E}\left[f_k\left(\theta_k^{t}\right)-f_k\left(\theta^{*}\right)\right]+\frac{1}{2} \eta_k^{2} L G
    \end{split}
\end{equation}
Eq. (\ref{eq:17}) can be rewritten as

\begin{equation}
\resizebox{1.0\hsize}{!}{$
        \mathbb{E}\left[f_k\left(\theta_k^{t+1}\right)-f_k\left(\theta_k^{*}\right)\right] \leq 
        \left(1-\mu \eta_k \sigma_k^{2}\right) \mathbb{E}\left[f_k\left(\theta_k^{t}\right)-f_k\left(\theta_k^{*}\right)\right] 
        +\frac{1}{2} \eta_k^{2} L G
$}
\end{equation}
This completes the proof of Theorem \ref{theorem: convergence}.
\end{proof}

\subsection{Security Analysis}
In EASTER, we use a blinding factor to hide the local embedding of passive parties.
If the blinding factor is security, we can ensure that other passive parties cannot infer the local features of passive parties. 
Next, we illustrate the security of the blinding factor.

In our model, each passive party’s private key \( SK_{l_k} \) is randomly generated within a cyclic group \( \mathbb{G} \) of prime order \( p \).
We compute each passive party's public key \( PK_{l_k} \) as $PK_{l_k} = g^{SK_{l_k}}$, where \(g\) is the generator of the group. 
The Computational Diffie–Hellman (CDH) problem ensures that given $PK_{l_k}$ and $g$, the probability of computing the $SK_{l_k}$ is negligible.
Therefore, even if an attacker obtains the public key \( PK_{l_k} \), deriving the private key \( SK_{l_k} \) in polynomial time remains infeasible, ensuring the security of the private key.

The blinding factor is generated based on the shared public key of the passive party. 
The shared public key \( CK_{k,j} = H((g^{s_{l_j}})^{s_{l_k}}) \) is derived from the public key of \( l_j \) and the private key of \( l_k \).   
The Computational Diffie–Hellman (CDH) problem ensures that given $PK_{l_j} = g^{SK_{l_j}}$ and $PK_{l_k} = g^{SK_{l_k}}$, the probability of computing the $CK_{k,j}$ is negligible.
Therefore, given access to all public key $PK_{l_k}=g^{s_{l_k}}$, the passive party $l_k$ is not able to infer the shared key $CK_{k,j}=H((g^{s_{l_j}})^{s_{l_k}})$, where $\forall k,j \in [1,K],j \neq k \neq K$, ensuring the security of the blinding factor.

Each passive party adds the blinding factor to its local embedding before sending it to the active party.   
We have already proven that the blinding factor is secure; even if an adversary possesses the public key, they are still unable to obtain the blinding factor. 
Therefore, the adversary cannot access the local embedding of the passive party, ensuring the security of the passive party's local data.  
Additionally, each passive party only has its private key, and even in the case of collusion between multiple passive parties, the shared key between the other passive parties cannot be obtained, and the true local embedding cannot be obtained.
Therefore, the EASTER method ensures that the passive party cannot obtain real local embedding with blinding factors from global embedding and thus cannot infer the raw features.

\section{Experiment Evaluation} \label{sec: exper}

\begin{table*}[!t]
\centering 
\caption{Performance (\%) comparison with baseline methods on benchmark datasets under heterogeneous settings.} 
\begin{tabular}{lcccc|cccc}
\toprule
\multirow{2}{*}{\textbf{Methods}} & \multicolumn{4}{c|}{\textbf{MNIST} } & \multicolumn{4}{c}{\textbf{FMNSIT}}  \\ \cline{2-9}
& $\theta_1$    & $\theta_2$ & $\theta_3$  & Avg & $\theta_1$ & $\theta_2$& $\theta_3$ & Avg \\

 \midrule
Local     &  25.60 $\pm$ 0.03     &  39.10 $\pm$ 0.04   & 35.62 $\pm$ 0.11 & 33.44 & 65.22 $\pm$ 0.16      &74.41 $\pm$ 0.07       &71.54 $\pm$ 0.12& 70.39\\
Pyvertical \cite{romanini2021pyvertical}             &97.15 $\pm$ 0.10       & 97.16 $\pm$ 0.09     & 97.21 $\pm$ 0.09  & 97.17 & 88.32 $\pm$ 0.20      & 88.11 $\pm$ 0.20     & 88.27 $\pm$ 0.19&88.23\\
C\_VFL \cite{castiglia2022compressed}       &90.91 $\pm$ 0.23       & 91.77 $\pm$ 0.30      & 90.87 $\pm$ 0.87 &91.18 &83.70 $\pm$ 0.20       & 82.10 $\pm$ 0.29    & 83.68 $\pm$ 0.27& 83.16\\
Agg\_VFL \cite{zhang2022adaptive}          &96.01 $\pm$ 0.09      & 94.34 $\pm$ 0.04       & 94.32 $\pm$ 0.06 &94.89& 86.06 $\pm$ 0.32      & 87.73 $\pm$ 0.33     & 89.20 $\pm$ 0.13 &87.66\\
\textbf{EASTER(ours)}  & \textbf{98.23 $\pm$ 0.06}      &  \textbf{97.98 $\pm$ 0.07}    & \textbf{98.27 $\pm$ 0.05}    & \textbf{98.16}&\textbf{89.71  $\pm$ 0.07}      &  \textbf{89.91 $\pm$ 0.07}    & \textbf{90.04 $\pm$ 0.02} &\textbf{89.88}\\
\midrule
\midrule
\multirow{2}{*}{\textbf{Methods}} & \multicolumn{4}{c|}{\textbf{CIFAR-10}} & \multicolumn{4}{c}{\textbf{CIFAR-100}}  \\ \cline{2-9}
& $\theta_1$    & $\theta_2$ & $\theta_3$  &Avg& $\theta_1$ & $\theta_2$& $\theta_3$ & Avg \\
\midrule
Local      & 56.20 $\pm$ 0.11 & 54.73 $\pm$ 0.35  & 53.16 $\pm$ 0.36  & 54.69& 23.86 $\pm$ 0.35  &  22.03 $\pm$ 0.15    & 23.36 $\pm$ 0.59 & 23.09\\
Pyvertical \cite{romanini2021pyvertical}           & 63.54 $\pm$ 0.39 & 62.53 $\pm$ 0.17 & 61.34 $\pm$ 0.28 & 62.47& 50.80 $\pm$ 0.78      &   50.54 $\pm$ 0.11    & 50.46 $\pm$ 0.08&50.36\\
C\_VFL \cite{castiglia2022compressed}          & 64.92 $\pm$ 0.26 & 64.40 $\pm$ 0.44 & 61.58 $\pm$ 0.49 & 63.63&  46.28 $\pm$ 0.37     &   46.65 $\pm$ 0.51     & 45.59 $\pm$ 0.78& 46.17\\
Agg\_VFL  \cite{zhang2022adaptive}            &74.52 $\pm$ 0.04 &74.09 $\pm$ 0.04 & 73.57 $\pm$ 0.10  &74.06 & 47.66 $\pm$ 0.09     &  48.65 $\pm$ 0.12    &  47.07 $\pm$ 0.06 &47.79\\
\textbf{EASTER(ours)}           & \textbf{79.80 $\pm$ 0.07}     &   \textbf{79.44 $\pm$ 0.08}    & \textbf{79.59 $\pm$ 0.10}   & \textbf{79.61}& \textbf{52.27 $\pm$ 0.11} &   \textbf{52.25 $\pm$ 0.06} & \textbf{51.76 $\pm$ 0.10} &\textbf{52.09}\\
\midrule
\midrule
\multirow{2}{*}{\textbf{Methods}} & \multicolumn{4}{c|}{\textbf{CINIC-10}} & \multicolumn{4}{c}{\textbf{CRITEO}}  \\ \cline{2-9}
& $\theta_1$    & $\theta_2$ & $\theta_3$ & Avg& $\theta_1$ & $\theta_2$& $\theta_3$  & Avg\\
\midrule
 Local        & 45.32 $\pm$ 0.17      &  45.44 $\pm$ 0.26    & 44.46 $\pm$ 0.24 & 45.07& 63.44 $\pm$ 0.48     &  67.92 $\pm$ 1.20    & 64.28 $\pm$ 0.49 &65.21\\
Pyvertical \cite{romanini2021pyvertical}          & 61.78 $\pm$ 0.71    &   62.29 $\pm$ 0.19    & 62.13 $\pm$ 0.32 & 62.07 & 74.56 $\pm$ 0.45     &   75.65 $\pm$ 0.58    & 72.34 $\pm$ 0.40 &74.18\\
C\_VFL \cite{castiglia2022compressed}   & 66.20 $\pm$ 0.55     &   65.37 $\pm$ 0.52    & 67.76 $\pm$ 0.46 & 66.44&  73.34 $\pm$ 1.11      &   75.57 $\pm$ 0.70     & 74.28 $\pm$ 0.32  &74.39\\
Agg\_VFL  \cite{zhang2022adaptive}       & 73.74 $\pm$ 0.40     &   73.79 $\pm$ 0.04    & 73.01 $\pm$ 0.05 &73.51& 75.76 $\pm$ 0.48     &   75.83 $\pm$ 0.31    & 74.71 $\pm$ 0.66&75.43\\
\textbf{EASTER(ours)}  & \textbf{77.37 $\pm$ 0.19}      &    \textbf{77.39 $\pm$ 0.06}     &  \textbf{76.74 $\pm$ 0.02}   &\textbf{77.16}& \textbf{77.14 $\pm$ 0.41}     &   \textbf{77.23 $\pm$ 0.31}    & \textbf{77.25 $\pm$ 0.83}&\textbf{77.21}\\
\bottomrule
\end{tabular}
\label{table: heter} 
\end{table*}

\subsection{Experimental Setup} \label{subsubsec: dataset}

\begin{table}[!t]
\centering
\caption{Summary of Benchmark Datasets. `num.' denotes `numerical' and `cate.' denotes `categorical'}
\label{table: datasets}
\begin{tabular}{lcccl}
\toprule
\textbf{Datasets}   & \textbf{\#Samples} & \textbf{\#Class} & \textbf{\#Feature} & \textbf{Type} \\ 
\midrule
MNIST      & 70,000      & 10     & 784 (28×28×1)     & Image \\ 
FMNIST     & 70,000      & 10     & 784 (28×28×1)     & Image \\ 
CIFAR-10    & 60,000     & 10     & 3,072 (32×32×3)   & Image \\ 
CIFAR-100   & 60,000     & 100    & 3,072 (32×32×3)   & Image \\ 
CINIC-10    & 270,000    & 10     & 3,072 (32×32×3)   & Image \\ 
CRITEO      & 45,000,000 & 2      & 39 (13num.+26cate.) & Tabular \\ 
\bottomrule
\end{tabular}
\end{table}

\subsubsection{Datasets}
We evaluate the EASTER framework on six benchmark datasets: MNIST, Fashion-MNIST (FMNIST), CIFAR-10, CINIC-10, CIFAR-100, and CRITEO, following prior works \cite{romanini2021pyvertical,castiglia2022compressed,zouvflair,ye2024vertical}.
We have summary benchmark datasets in Table \ref{table: datasets}.
\textbf{(1) MNIST} dataset~\cite{lecun1998gradient}, consisting of grayscale handwritten digit images of size 28 $\times$ 28 $\times$ 1, contains 60,000 training samples and 10,000 test samples.
\textbf{FMNIST}~\cite{xiao2017fashion} is an image dataset consisting of grayscale images (28 $\times$ 28 $\times$ 1) of clothing items across 10 classes, comprising a total of 70,000 images, with 60,000 images for training and 10,000 images for testing.
\textbf{(3) CIFAR-10} \cite{krizhevsky2009learning} consists of 60,000 color images (32 $\times$ 32 $\times$ 3) distributed evenly across 10 classes, with 6,000 images per class. 
\textbf{(4) CIFAR-100} \cite{krizhevsky2009learning}  cotains the same number of images as CIFAR-10 but spans 100 classes, with 600 images per class, offering greater categorical diversity and increased complexity.
\textbf{(5) CINIC-10} \cite{NEURIPS2021_2f2b2656} is an extended version of the CIFAR-10 dataset, created by augmenting it with ImageNet images \cite{deng2009imagenet}. It contains 270,000 images across both training and testing sets, with 10 distinct classes. The training set of CINIC-10 is 3 times larger than that of CIFAR-10, providing a more robust dataset for performance evaluation.
\textbf{(6) CRITEO} \cite{10380676} is specifically designed for \textit{Click-Through Rate} (CTR) prediction tasks. It comprises over 45 million samples and includes a variety of 13 numerical and 26 categorical features. 
The dataset is characterized by its large scale, high dimensionality, and sparsity of features. 
For the experimental evaluation, the data are divided into training and testing sets, comprising 80\% and 20\% of the data, respectively.

\subsubsection{Models}
We selected different neural network architectures based on the characteristics of each dataset. 
Specially, for the MNIST and FMNIST datasets, we employed \textit{Multi-Layer Perceptrons} (MLP) \cite{taud2018multilayer}, \textit{Convolutional Neural Networks} (CNN) \cite{albawi2017understanding}, and LeNet \cite{lecun2015lenet}. 
For the CIFAR-10, CIFAR-100, and CINIC-10 datasets, ResNet18, ResNet34, and ResNet50 \cite{subaar2024investigating} were utilized. 
Due to the limited number of training samples per class in the CIFAR-100 dataset, pre-trained versions of ResNet-18, ResNet-34, and ResNet-50 were employed to enhance model performance. 
For the CRITEO dataset, models such as DeepFM \cite{guo2017deepfm}, Wide \& Deep \cite{cheng2016wide}, and CNN were selected.
In the heterogeneous model scenario, each participant selects a model that best suits their local requirements.
Furthermore, in the homogeneous model setting, all participants select the same model architecture for collaborative optimization.
During the training process, participants randomly selected a network based on local resources, and multiple participants formed heterogeneous models for collaborative training.

\subsubsection{Baselines}

To evaluate the performance of EASTER under heterogeneous models, we conducted a comparative analysis with baseline methodologies. 
The comparison included local training, where the active party trains a model separately using their local features, along with three recent VFL studies as follows.
\begin{itemize}
    \item Pyvertical \cite{romanini2021pyvertical} implemented basic SplitVFL model training and divided the entire model into a top model and a bottom model. The active party owned the top-level model, and the passive party owned the bottom-level model. The active party and the passive party collaboratively trained the global model.
    \item C\_VFL \cite{castiglia2022compressed} reduced the communication overhead of VFL by compressing the amount of transmitted data based on traditional SplitVFL.
    \item Agg\_VFL \cite{zhang2022adaptive} was an AggVFL method used to solve the imbalanced features of each participant.
\end{itemize}

\subsubsection{Implementation Details}
To ensure a fair comparison, all methods were evaluated under the same experimental settings.
Specifically, the EASTER and baseline methods were implemented using the PyTorch \cite{paszke2019pytorch} framework. 
In VFL, $C$ participants were employed, with one party assuming an active role and the remaining three parties adopting passive roles. 
The data set comprising all samples was partitioned into $C$ distinct portions vertically. 
We set $C$ = 4 to evaluate our method.
Additionally, we evaluated the training time and memory overheads of our method by varying the number of participants.
Furthermore, we set the batch size to 128. 
Furthermore, we set the batch size to 128 and used different learning rates for each dataset. 
Specifically, the learning rate was set to 0.01 for MNIST and FMNIST, 0.1 for CIFAR-10, CIFAR-100, and CINIC-10 (with exponential decay during training), and 0.001 for CRITEO. We set the embedding size to 128, and the number of layers in both the local embedding and predicting models was the same.

\begin{table*}[!t]
\centering 
\caption{Performance (\%) comparison with baseline methods on benchmark datasets under homogeneous settings.} 
\begin{tabular}{lcccc|cccc}
\toprule
\multirow{2}{*}{\textbf{Methods}} & \multicolumn{4}{c|}{\textbf{MNIST} } & \multicolumn{4}{c}{\textbf{FMNIST}}  \\ \cline{2-9}
& $\theta_1$    & $\theta_2$ & $\theta_3$ & Avg & $\theta_1$ & $\theta_2$& $\theta_3$ & Avg \\
 \midrule
Local     &  25.60 $\pm$ 0.03     &  39.10 $\pm$ 0.04   & 35.62 $\pm$ 0.11 & 33.44 & 65.55 $\pm$ 3.41       &76.24 $\pm$ 1.04       &70.40 $\pm$ 2.37 & 70.73\\
Pyvertical \cite{romanini2021pyvertical}    &92.85 $\pm$ 0.09 & 98.07 $\pm$ 0.03 &  95.59 $\pm$ 0.09 & 95.50 & 83.80 $\pm$ 0.08      & 89.15 $\pm$ 0.11     &85.31 $\pm$ 0.16 & 86.09\\
C\_VFL \cite{castiglia2022compressed}       &86.35 $\pm$ 0.12   & 95.94 $\pm$ 0.10      & 92.95 $\pm$ 0.18 & 91.75 & 81.78 $\pm$ 0.19      & 83.25 $\pm$ 0.27      & 81.23 $\pm$ 0.14 & 82.09\\
Agg\_VFL \cite{zhang2022adaptive}          &93.67 $\pm$ 0.05      & 97.68 $\pm$ 0.06       & 95.19 $\pm$ 0.04 & 95.51 & 86.55 $\pm$ 0.14      &  90.66 $\pm$ 0.37       & 86.23 $\pm$ 0.34 & 87.81\\
\textbf{EASTER(ours)}  & \textbf{96.39 $\pm$ 0.03} & \textbf{96.79 $\pm$ 0.09} &  \textbf{97.95 $\pm$ 0.06} & \textbf{97.04} & \textbf{87.44 $\pm$ 0.08}  & \textbf{91.34 $\pm$ 0.07}  & \textbf{87.47 $\pm$ 0.09} & \textbf{88.75}\\
\midrule

\midrule
\multirow{2}{*}{\textbf{Methods}} & \multicolumn{4}{c|}{\textbf{CIFAR-10} } & \multicolumn{4}{c}{\textbf{CIFAR-100}}  \\ \cline{2-9}
& $\theta_1$    & $\theta_2$ & $\theta_3$ & Avg & $\theta_1$ & $\theta_2$& $\theta_3$ & Avg \\
\midrule
Local      & 56.20 $\pm$ 0.11 & 54.73 $\pm$ 0.35  & 53.16 $\pm$ 0.36  & 54.70 & 23.86 $\pm$ 0.35  &  22.03 $\pm$ 0.15    & 23.36 $\pm$ 0.59 & 23.08\\
Pyvertical \cite{romanini2021pyvertical}           & 61.18 $\pm$ 0.67 & 62.87 $\pm$ 0.37 & 61.10 $\pm$ 0.15 & 61.72 & 32.07 $\pm$ 0.18  & 35.82 $\pm$ 0.66 &  36.41 $\pm$ 0.22 & 34.77\\
C\_VFL \cite{castiglia2022compressed}          & 61.83 $\pm$ 0.43 & 61.52 $\pm$ 0.31 & 61.82 $\pm$ 0.38  & 61.71 & 33.07 $\pm$ 0.42  & 34.63 $\pm$ 0.29 &  35.51 $\pm$ 0.23  & 34.40\\
Agg\_VFL  \cite{zhang2022adaptive}            &77.33 $\pm$ 0.10 & 76.47 $\pm$ 0.13 & 69.00 $\pm$ 0.27  & 74.27 & 42.29 $\pm$ 0.08 & 45.45 $\pm$ 0.05 &  45.22 $\pm$ 0.07 & 44.32\\
\textbf{EASTER(ours)}           & \textbf{79.86 $\pm$ 0.10} & \textbf{78.01 $\pm$ 0.11} &  \textbf{78.04 $\pm$ 0.13} & \textbf{78.64} & \textbf{51.67 $\pm$ 0.11} &  \textbf{52.98 $\pm$ 0.13} & \textbf{51.28 $\pm$ 0.07} & \textbf{51.98}\\
\midrule
\midrule
\multirow{2}{*}{\textbf{Methods}} & \multicolumn{4}{c|}{\textbf{CINIC-10}} & \multicolumn{4}{c}{\textbf{CRITEO}}  \\ \cline{2-9}
& $\theta_1$    & $\theta_2$ & $\theta_3$ & Avg & $\theta_1$ & $\theta_2$& $\theta_3$ & Avg \\
\midrule
 Local        & 45.32 $\pm$ 0.17      &  45.44 $\pm$ 0.26    & 44.46 $\pm$ 0.24 & 45.07 & 63.44 $\pm$ 0.48     &  67.92 $\pm$ 1.20    & 64.28 $\pm$ 0.49 & 65.21\\
Pyvertical \cite{romanini2021pyvertical}          & 60.38 $\pm$ 0.93 & 58.62 $\pm$ 0.86 &  58.78 $\pm$ 0.82 & 59.26 &71.25 $\pm$ 0.41 & 72.58 $\pm$ 0.44 &  72.59 $\pm$ 0.48 & 72.14\\
C\_VFL \cite{castiglia2022compressed}   & 66.44 $\pm$ 0.61 & 65.59 $\pm$ 0.57 &  67.83 $\pm$ 0.68 & 66.62 &  67.29 $\pm$ 1.92 & 72.39 $\pm$ 0.31 &  68.61 $\pm$ 0.49  & 69.43\\
Agg\_VFL  \cite{zhang2022adaptive}       & 73.74 $\pm$ 0.40     &   73.79 $\pm$ 0.04    & 73.01 $\pm$ 0.05 & 73.51 &74.37 $\pm$ 0.60 & 77.07 $\pm$ 0.76 &  76.08 $\pm$ 0.41 & 75.84\\
\textbf{EASTER(ours)}  & \textbf{73.87 $\pm$  0.65} & \textbf{75.96 $\pm$ 0.16} &  \textbf{73.90 $\pm$ 0.31} & \textbf{74.58} & \textbf{77.04 $\pm$ 0.48} & \textbf{77.37 $\pm$ 0.91} &  \textbf{77.41 $\pm$ 0.58} & \textbf{77.27}\\
\bottomrule
\end{tabular}
\label{table: homo} 
\end{table*}

\subsection{Evaluation of Model Accuracy}
\subsubsection{Heterogeneous Model VFL Methods}
We conducted extensive experiments to evaluate the accuracy performance of EASTER and various baseline methods.
To ensure a fair comparison, we adapted the baseline implementations by replacing the homogeneous local models across clients with heterogeneous ones (i.e., each passive party was assigned a different local model during training).
We evaluated the accuracy of EASTER on multiple datasets under heterogeneous models, and the experimental results are shown in Table \ref{table: heter}.
From Table \ref{table: heter}, we found that compared with \textit{Local} method, the model accuracy of EASTER had been significantly improved. 
This was because the \textit{Local} method only used some of the local features of the active party to train the model, while EASTER collaborated with all the features of the four participants to train the model. 
Therefore, as the features of the training data set increased, the training accuracy of the model continued to improve. 
Compared with the better method Agg\_VFL, the model accuracy of EASTER was improved by 7.22\% under the CIFAR-10 dataset.
This was because the intermediate results or prediction results aggregated by the existing VFL method had more local model information, while the embedding aggregated by EASTER had less local model information. 
Local model information had a certain negative impact on the training results.
Therefore, compared with existing methods, the accuracy of the heterogeneous model trained by EASTER was better.
Furthermore, EASTER assigns each participant an independent heterogeneous model, enabling the simultaneous training of multiple heterogeneous local models. 
This architectural design enhances model diversity and contributes to improved overall accuracy.

\subsubsection{Homogeneous Model VFL Methods}
We evaluate the test accuracy of our proposed method, EASTER, and compare it with several baseline approaches under homogeneous model settings across a variety of datasets. The detailed results are presented in Table \ref{table: homo}.
As shown in the table, EASTER consistently achieves superior performance compared to the baselines. For example, EASTER outperforms the strongest baseline, Agg\_VFL, by approximately 2\% in test accuracy on the CIFAR-10 dataset.
This improvement stems from the structural limitations of baseline methods, where each participant either holds only a partial model or relies on aggregated predictions that are not directly trainable. As a result, these approaches fail to capture the complete local feature representations during training, leading to suboptimal global performance.
In contrast, EASTER assigns each participant a fully independent local model, and its embedding aggregation mechanism allows for the integration of representative local features. This design enables more effective model optimization at each client, thereby enhancing overall accuracy.
Therefore, EASTER consistently exhibits superior performance relative to existing methods across both homogeneous and heterogeneous local model.

\begin{table*}[!t]
\centering
\caption{Comparison of test accuracy (ACC), area under the curve (AUC), time cost (TC), and space cost (SC) of EASTER and baseline methods on the CINIC-10 and CRITEO datasets with varying numbers of clients. }  
\begin{tabular}{c|cc|cc|cc|cc}
\toprule
 \multirow{2}{*}{\textbf{Methods}} & \multicolumn{2}{c|}{\textbf{FMNIST} } & \multicolumn{2}{c|}{\textbf{CIFAR-100}}& \multicolumn{2}{c|}{\textbf{CINIC-10} } & \multicolumn{2}{c}{\textbf{CRITEO}}\\ \cmidrule{2-9}  
& ACC (\%) $\color{blue}{\uparrow}$ & Com (MB) $\color{blue}{\downarrow}$ & ACC (\%) $\color{blue}{\uparrow}$  & Com (MB) $\color{blue}{\downarrow}$ &  ACC (\%) $\color{blue}{\uparrow}$ & Com (MB) $\color{blue}{\downarrow}$  & AUC (\%) $\color{blue}{\uparrow}$ & Com (MB) $\color{blue}{\downarrow}$
\\ \hline
Pyvertical& 88.11 & 1401.62  &50.54  &262500.01 & 62.29 & 324843.75   & 75.65 &   11022.95     \\
C\_VFL    & 82.10 & 1602.17  &46.65 &284375.80  & 65.37  & 354375.31   &75.57  & 24801.63            \\
Agg\_VFL  & 87.73  & 137.43  &48.65  &686.71  &73.79   & 412.31     &75.83  & 184.53       \\
EASTER   & \textbf{89.91}   & \underline{840.45}   & \textbf{52.25}    & \underline{206262.58}   & \textbf{77.39}  &\underline{278482.81} & \textbf{77.37}   & \underline{7113.64}  \\
 \bottomrule
\end{tabular}
\label{table: table_comm} 
\end{table*}

\begin{figure*}[!t] 
	\begin{minipage}{0.24\linewidth}
		\centerline{\includegraphics[width=\textwidth]{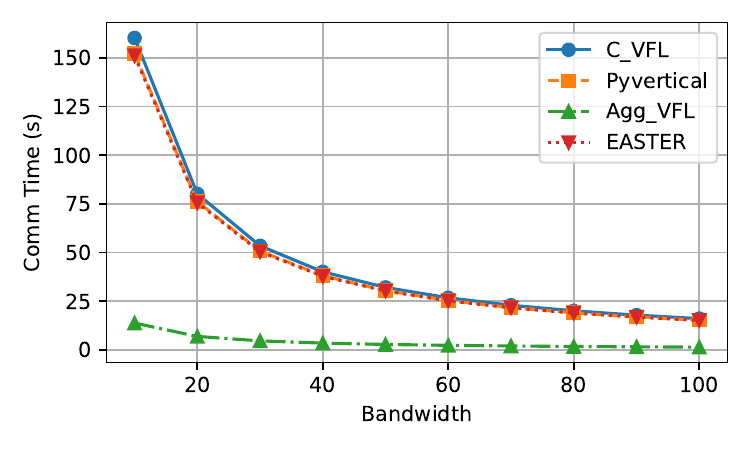}}
		\centerline{(a) FMNIST}
	\end{minipage}
    \begin{minipage}{0.24\linewidth}
		\centerline{\includegraphics[width=\textwidth]{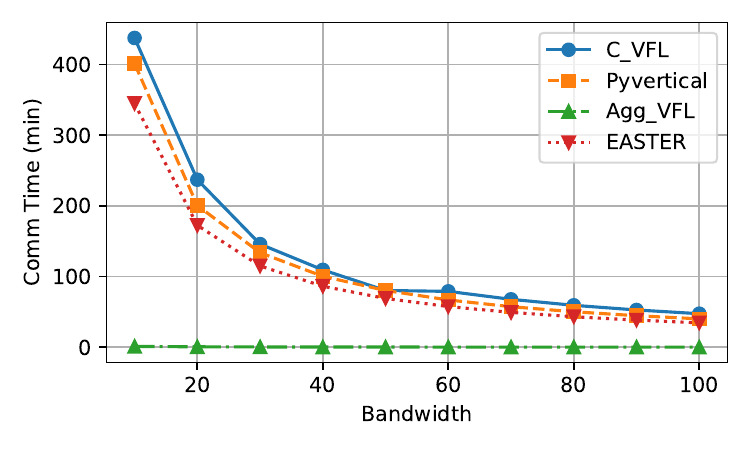}}
		\centerline{(b) CIFAR-100}
	\end{minipage}
	\begin{minipage}{0.24\linewidth}
		\centerline{\includegraphics[width=\textwidth]{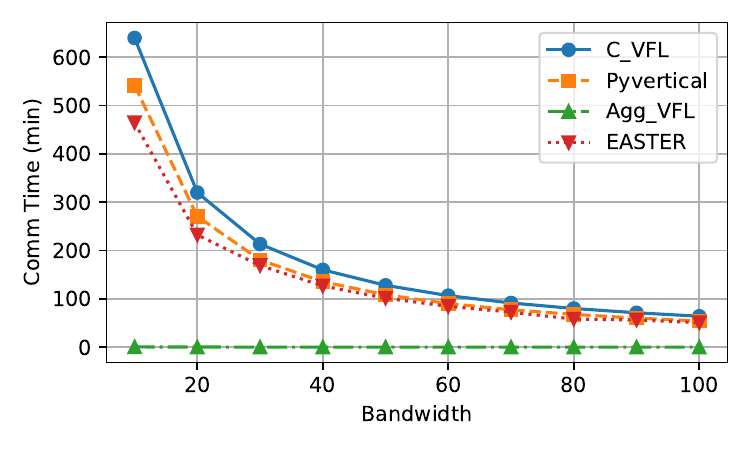}}
		\centerline{(c) CINIC-10}
	\end{minipage}
    \begin{minipage}{0.24\linewidth}
		\centerline{\includegraphics[width=\textwidth]{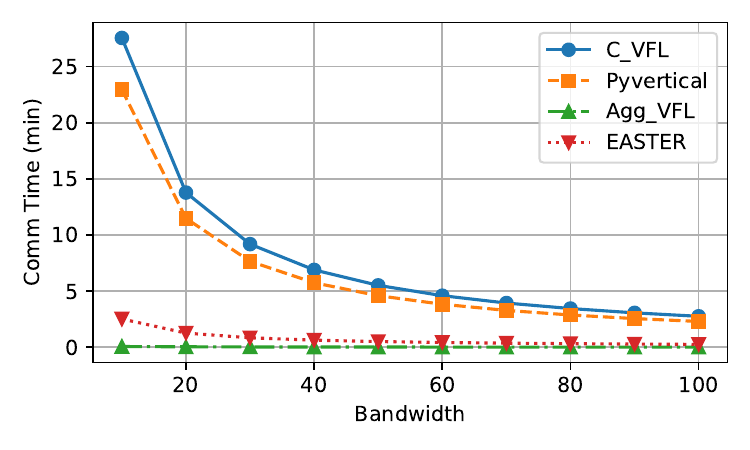}}
		\centerline{(d) CRITEO}
	\end{minipage}
\caption{Communication time overhead for model convergence across four datasets at different network bandwidths.}
\label{Fig: bandwidth}
\end{figure*}

\begin{figure*}[!t] 
	\begin{minipage}{0.24\linewidth}
		\centerline{\includegraphics[width=\textwidth]{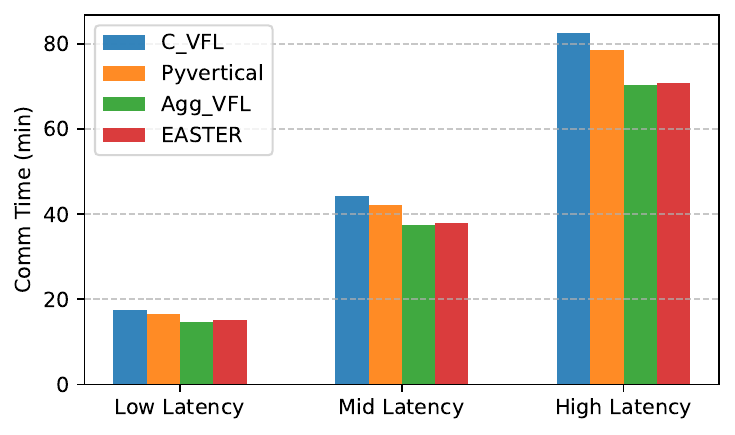}}
		\centerline{(a) FMNIST}
	\end{minipage}
    \begin{minipage}{0.24\linewidth}
		\centerline{\includegraphics[width=\textwidth]{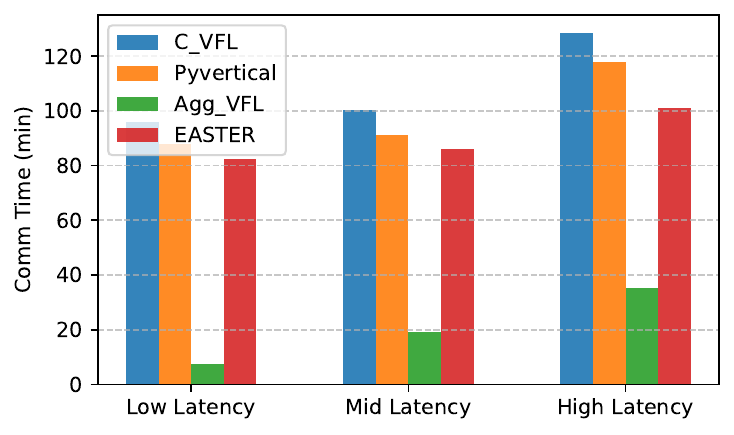}}
		\centerline{(b) CIFAR-100}
	\end{minipage}
	\begin{minipage}{0.24\linewidth}
		\centerline{\includegraphics[width=\textwidth]{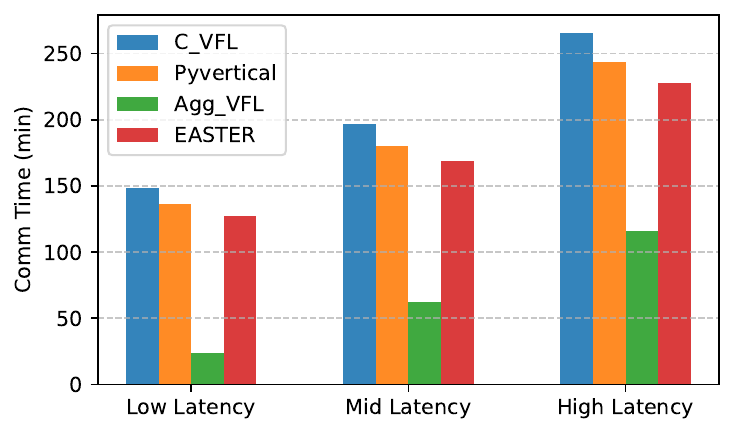}}
		\centerline{(c) CINIC-10}
	\end{minipage}
    \begin{minipage}{0.24\linewidth}
		\centerline{\includegraphics[width=\textwidth]{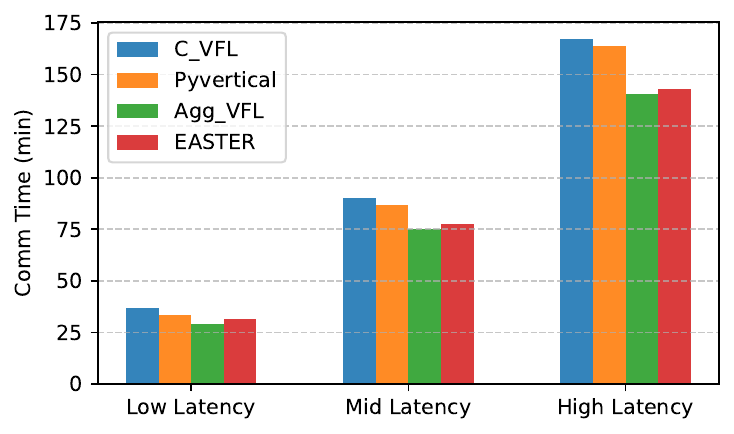}}
		\centerline{(d) CRITEO}
	\end{minipage}
\caption{Communication time overhead for model convergence across four datasets at different network latency.}
\label{Fig: latency}
\end{figure*}

\subsection{Evaluation of Communication Overhead}
We evaluated the communication overhead of our method and compared it to baseline approaches across four datasets.
The evaluation focused on three primary aspects: communication volume, communication time under varying bandwidth conditions, and communication time under different network latencies.
Communication volume is presented in Table \ref{table: table_comm}, where our method achieves the highest model accuracy at convergence, despite its marginally higher communication volume compared to Agg\_VFL.
For example, on the CIFAR-100 dataset, our method reaches an accuracy of 52.27\%, surpassing the highest accuracy achieved by Pyvertical by 3.42\% and Agg\_VFL by 7.44\%.
The communication volume for our method is 206,262.58 MB, 27\% lower than Pyvertical.

Regarding communication time, Fig. \ref{Fig: bandwidth} shows that as bandwidth increases, the time required for model convergence decreases.
Fig. \ref{Fig: latency} illustrates the effect of network latency (low latency: 1 ms–30 ms, medium latency: 30 ms–50 ms, high latency: 50 ms–100 ms) on communication time, with bandwidth fixed at 50 Mbps.
The results indicate that communication time increases as network latency rises.
For FMNIST and CRITEO, where data exchanged per round is small, communication time is more sensitive to latency, leading to minimal differences across the methods. In contrast, for datasets like CIFAR-100 and CINIC-10, which involve larger data exchanges, communication time is influenced by both bandwidth and latency for all methods except Agg\_VFL.
Our method transmits embedding values, which are larger in size compared to prediction values transmitted by Agg\_VFL, resulting in higher communication volume but superior accuracy.
In summary, our method achieves the highest accuracy at model convergence with reasonable communication volume and time overhead.
\subsection{Evaluation of Time and Memory Overhead} \label{subsec: ETMO}
We evaluated the time and memory overhead of our method on the CINIC-10 and CRITEO datasets, with the experimental results presented in Table \ref{table: multi_clients}. 
In this experiment, we set the total number of participants to 10, meaning that the dataset’s features were split along the feature dimension into 10 equal parts, with each client having one part of the data features and a heterogeneous model.
We assessed the accuracy or AUC, training time, and memory overhead during collaborative model training with varying numbers of participants.

As shown in Table \ref{table: multi_clients}, we observe that, for both the CINIC-10 and CRITEO datasets, our method achieves higher accuracy at convergence compared to the baseline method.
Specifically, for the CINIC-10 dataset, the baseline method reaches lower accuracy, resulting in lower training complexity and hence lower time and memory overheads. However, our method achieves optimal model accuracy with acceptable time and memory overhead.
On the other hand, as the number of clients increases, both the workload and communication among participants grow.
This leads to improvements in model accuracy, training time, and memory usage.
For instance, on the CINIC-10 dataset, when there are 2 and 10 clients, our method’s model accuracy improves from 54.22\% to 72.79\%, representing an improvement of 34.24\%. 
The training time increases from 34.58 minutes to 348.51 minutes, referring to the model optimization time, not inference time, and thus remains within an acceptable range. 
Additionally, the memory overhead increases from 0.49GB to 4.22GB, which can be handled by most standard servers.
The findings demonstrated that EASTER was appropriate for training heterogeneous models as local embedding has less information.

\begin{table*}[!t]
\centering
\caption{Comparison of test accuracy (ACC), area under curve (AUC), time cost (TC), and memory cost (MC) of EASTER and baseline methods on the CINIC-10 and CRITEO datasets with varying numbers of clients. ACC for the CINIC dataset, while AUC for the CRITEO dataset. The memory cost (SC) is measured in gigabytes (GB) for CINIC-10 and in megabytes (MB) for CRITEO datasets.}  
\begin{tabular}{c|cccccc|ccccc}
\toprule
\multirow{2}{*}{} & \multirow{2}{*}{\textbf{Methods}} & \multicolumn{5}{c|}{\textbf{CINIC-10} \textit{\{ACC (\%) / TC (min) / SC (GB) \}} } & \multicolumn{5}{c}{\textbf{CRITEO} \textit{\{AUC (\%) / TC (min) / SC (MB) \}}}\\ \cmidrule{3-12}  
& & C = 2 & C = 4 & C = 6 & C = 8 & C = 10 & C = 2    & C = 4 & C = 6 & C = 8 & C = 10
\\ \hline
{\multirow{4}{*}{ACC / AUC (\%) \color{blue}{$\uparrow$}}} 
& Pyvertical& 49.98 & 58.41  &63.67  &66.23  & 67.40 & 69.45   & 72.35 &   72.63    &73.30  &74.36  \\
& C\_VFL   & 50.17 & 58.66  &63.45  &65.37  & 65.91 & 68.34   &71.59  & 72.54      &73.32  &74.08       \\
& Agg\_VFL & 53.17 & 60.36  &66.12  &67.25  &67.66  & 65.54   &71.98  & 72.12      &73.47  & 74.12  \\
 & EASTER(ours)& \textbf{54.22}  &\textbf{62.57}   & \textbf{67.50} & \textbf{72.33} & \textbf{72.79} &\textbf{70.40} & \textbf{74.52}   & \textbf{75.51} & \textbf{75.93} &  \textbf{76.63} \\
\midrule
\multirow{4}{*}{TC (min) \color{blue}{$\downarrow$}}
& Pyvertical &23.26  & 31.77  &45.13  &57.70  & 72.62 &59.23   &62.33  & 91.30      & 121.00 & 144.81 \\
& C\_VFL &47.94  & 84.14  & 86.98 &115.33  &157.82  &60.87   &114.98  &182.47       & 209.08 &217.16  \\
& Agg\_VFL  & 30.54 & 171.14  &313.07  & 534.75 &810.34  &19.86   &22.82  & 42.70      & 57.94 & 76.09  \\
& EASTER(ours)  &34.58  & 133.50  &229.78  &263.95  & 348.51 &\underline{23.45} &\underline{31.19}    & \underline{57.57} & \underline{69.24}      &\underline{98.26}   \\
\midrule
\multirow{5}{*}{MC (GB / MB) \color{blue}{$\downarrow$}}                   
&Pyvertical  &0.34  &0.48   &0.62  &0.74  &0.92  & 25.01   & 115.75 &  117.11     &153.29  &  195.02  \\
&C\_VFL    &0.34  & 0.49  &0.63  &0.75  &0.93  & 25.07  & 116.33 & 117.55      & 155.97 & 197.85 \\
&Agg\_VFL  &0.56  &1.54   &2.27  &3.03  &4.23  &49.22   & 65.33 & 91.86     & 93.92 & 126.85  \\
&EASTER(ours) &0.49  & 1.45  &2.09  &2.88  &4.22  &\textbf{49.10} & \textbf{65.31}   & \textbf{91.66} & \textbf{93.55}      &\textbf{126.37}   \\
\bottomrule
\end{tabular}
\label{table: multi_clients} 
\end{table*}

\begin{table}[!t]
\caption{Comparison test accuracy and time of EASTER on heterogeneous devices}
\resizebox{0.5 \textwidth}{!}{
\begin{tabular}{ccc|cccc|cccc}
\toprule
&   &   & \multicolumn{4}{c|}{MNIST}&\multicolumn{4}{c}{FMNIST} \\
  \midrule
$\mathcal{A}$ & $\mathcal{B}$ & $\mathcal{C}$ &   $\theta_1$ (\%)     & $\theta_2$ (\%)       &  $\theta_3$ (\%)  & Time (min)  &   $\theta_1$ (\%)     & $\theta_2$ (\%)       &  $\theta_3$ (\%) & Time (min)  \\
\midrule
1 & 1 & 1 & 98.23  & 97.98    & 98.27  & 7.18 & 89.71     & 89.91    &90.04   &  9.99 \\
1 & 1 & 0 & 97.91   & 97.90   &  97.89  &20.94 & 88.71   & 88.73   & 88.71    & 15.17   \\
1 & 0 & 0 & 97.94   & 97.93  &97.91    &26.56  & 88.97   & 89.01  & 88.95   & 21.47  \\
0 & 0 & 0 & 97.85   & 97.84   &97.85 &35.87  & 88.85   & 88.76   & 88.79   & 28.86  \\
\midrule
\midrule
&   &   & \multicolumn{4}{c|}{CIFAR10}&\multicolumn{4}{c}{CIFAR100} \\
  \midrule
$\mathcal{A}$ & $\mathcal{B}$ & $\mathcal{C}$ &    $\theta_1$ (\%)     & $\theta_2$ (\%)       &  $\theta_3$ (\%)  & Time (min)  &   $\theta_1$ (\%)     & $\theta_2$ (\%)       &  $\theta_3$ (\%)  & Time (min)  \\
\midrule
1 & 1 & 1 &79.80  & 79.44 & 79.59   &32.58 & 52.77   & 52.27   & 52.25   & 49.12   \\
1 & 1 & 0 & 80.85 & 80.49 & 80.31   &43.52 & 52.62   & 52.47   &  52.64  & 73.04  \\
1 & 0 & 0 & 80.21   & 80.49 &80.35  &63.03  &    51.06 & 51.36   &50.87    & 79.60  \\
0 & 0 & 0 & 79.97   & 79.69  &79.64 &70.91  &    53.25 & 54.17  &  53.90  & 107.80 \\
\midrule
\midrule
  &   &   & \multicolumn{4}{c|}{CINIC10}&\multicolumn{4}{c}{CRITEO} \\
  \midrule
$\mathcal{A}$ & $\mathcal{B}$ & $\mathcal{C}$ &  $\theta_1$ (\%)       & $\theta_2$ (\%)       &  $\theta_3$ (\%)  & Time (min)  &   $\theta_1$ (\%)     & $\theta_2$ (\%)        &  $\theta_3$ (\%)  & Time (min)  \\
\midrule
1 & 1 & 1 &  77.37 & 77.39 & 76.74   & 74.00   & 77.14    &   77.23    & 77.25   &40.39   \\
1 & 1 & 0 &  77.64 & 77.53 & 77.08   & 151.51  & 77.23   & 77.24    & 77.28   &72.47   \\
1 & 0 & 0  & 77.41 & 77.49 & 76.88   & 187.86  & 77.21   & 77.24    & 77.32   &100.33 \\
0 & 0 & 0 & 77.22  & 77.30 & 76.80   & 258.97  & 77.20   & 77.21    & 77.31   &127.34  \\
\bottomrule
\end{tabular}
}
\label{table: heter_device} 
\end{table}

\subsection{Evaluation on Heterogeneous Party}

We evaluate the performance of our method with heterogeneous participants, focusing on heterogeneous devices and model parameters. First, we assess the model's test accuracy and training time across different numbers of high-performance devices (high bandwidth, low latency) and low-performance devices (low bandwidth, high latency). The results are presented in Table \ref{table: heter_device}. In this table, $\mathcal{A}$, $\mathcal{B}$, and $\mathcal{C}$ represent the client devices participating in training, where “1” indicates a high-performance device, and “0” indicates a low-performance device. We evaluated training with 0, 1, 2, and 3 low-performance devices across six datasets. The results show that as the number of low-performance devices increases, the model’s accuracy remains stable, while the training time increases progressively. For example, on the CINIC dataset, the model’s accuracy is around 77\%. With all three devices as high-performance, training time is 74 minutes, while with all three as low-performance, it increases to 258.97 minutes. The additional training time is within tolerable limits. Therefore, our method maintains strong performance even with heterogeneous devices.

\begin{figure}[!t] 
    \begin{minipage}{0.49\linewidth}
		\centerline{\includegraphics[width=\textwidth]{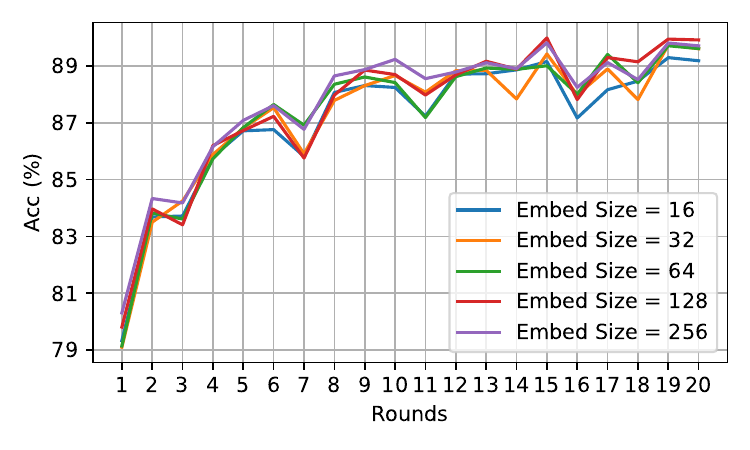}}
		\centerline{(a) Embedding Sizes}
	\end{minipage}
	\begin{minipage}{0.49\linewidth}
		\centerline{\includegraphics[width=\textwidth]{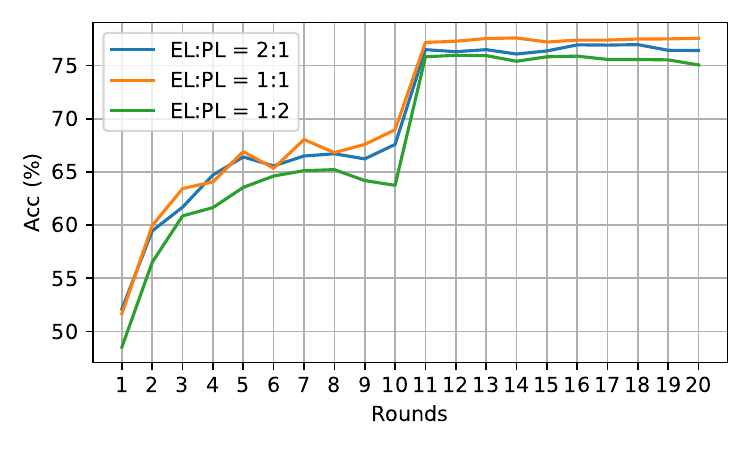}}
		\centerline{(b) Local Model Layers}
	\end{minipage}
\caption{{Test accuracy of our method under different embedding sizes and local model architectures. ``EL:PL'' is the ratio of the number of layers in the local embedding model and the local prediction model.}}
\label{Fig: Embedding_layers}
\end{figure}

Next, we evaluated the impact of local embedding size and the ratio of local embedding model layers (EL) to prediction model layers (PL) on model accuracy. The experimental results are presented in Fig. \ref{Fig: Embedding_layers}. Fig. \ref{Fig: Embedding_layers}(a) illustrates the model accuracy of EASTER with embedding sizes of 16, 32, 64, 128, and 256 on the FMNIST dataset. As shown in Fig. \ref{Fig: Embedding_layers}(a), the model achieves its highest accuracy when the batch size is set to 128. This is because smaller embeddings fail to adequately represent the local features, while larger embeddings tend to exhibit poorer generalization performance. 
Fig. \ref{Fig: Embedding_layers}(b) presents the model accuracy for different ratios of local embedding layers (EL) to prediction layers (PL), specifically 2:1, 1:1, and 1:2. The results indicate that the model achieves optimal performance when EL:PL is 1:1. 
This suggests that extreme ratios of embedding to prediction layers are less effective at capturing the local data features, leading to a decrease in model accuracy. Therefore, selecting the appropriate embedding size and layer configuration is crucial for optimizing model performance.
Therefore, selecting the appropriate embedding size and layer configuration is crucial for optimizing model performance.

\section{Discussion} \label{sec: dis}
We propose a privacy-preserving approach for heterogeneous model optimization in VFL, and both theoretical analysis and experimental results indicate that, compared to existing baseline methods, EASTER demonstrates superior performance.
However, there are still practical considerations, limitations, and opportunities for future research in the deployment of EASTER in real-world applications. 
In this section, we outline the practical deployment of EASTER and future research directions.

\textbf{Applicability.}
EASTER can be deployed in common application domains of vertical federated learning (VFL), such as recommendation systems \cite{huang2023incentive}, finance \cite{chatterjee2023federated}, healthcare \cite{nguyen2022federated}, and others.
It is important to note that, in practical applications, EASTER focuses on enabling privacy-preserving, multi-party heterogeneous model optimization in scenarios where devices already possess local data, without involving the data collection process.
For example, we propose the following approach to deploying EASTER in ISCC-based multi-device edge AI inference system \cite{wen2023task}. Since EASTER is designed to support collaborative training when devices already have local data, in ISCC-based multi-device edge AI inference tasks, we generate and process data for each edge node using the edge wireless sensing simulator described in the work on ISCC-based multi-device edge AI. Additionally, we configure heterogeneous model architectures and consistent embeddings for each edge node, ensuring that the data remains within the local domain to enable distributed model training, while employing privacy-preserving techniques for model optimization. Furthermore, during the optimization process, we apply the resource allocation algorithm from the work on multi-device edge AI to allocate resources across edge devices, resulting in an optimized, privacy-preserving multi-heterogeneous inference model, which is then deployed to edge devices for inference.

\textbf{Future Direction.}
Our method can be extended to large-scale datasets and multi-party scenarios, as demonstrated by the experimental results in Section \ref{subsec: ETMO}. Although our method is scalable to large-scale datasets and multi-party scenarios, as the dataset size and number of participants increase, the model's test accuracy continues to improve, while the communication and memory overheads continue to rise. This is due to varying performance levels among participants, which cause time delays in computing embeddings, and frequent data exchanges are required during the training process. Therefore, future research focusing on asynchronous training in heterogeneous vertical federated learning and reducing communication overhead will be of practical value for enabling the widespread adoption of VFL.
At the same time, EASTER effectively implements privacy-preserving multi-party heterogeneous model optimization, but the current research focuses on unimodal datasets. The design's emphasis on unimodal feature alignment limits its applicability in multimodal WoT scenarios, where devices generate diverse data streams, such as sensor signals, images, and audio.
In the future, integrating large multimodal foundation models (e.g., CLIP) for cross-modal feature fusion while preserving VFL’s privacy guarantees will be crucial.
These improvements aim to expand EASTER’s applicability in real-world WoT deployments that require seamless multimodal interoperability.

\section{Related Work} \label{sec: related}
\subsection{Heterogeneous Model in Federated Learning}
The heterogeneity model in FL refers to varied structures of participants' local model and heterogeneity is deemed to be one of the key challenges in FL \cite{hu2021mhat, huang2022learn}.
Recent research mostly can be grouped into two types, including Knowledge Distillation (KD)-based methods \cite{zhu2021data,cho2023communication,zhang2023towards,he2020group} and Partial Training (PT)-based methods \cite{alam2022fedrolex}.
A typical KD-based method extracts knowledge from a teacher model into student models with varied architectures.
FedGKT \cite{he2020group} is a group knowledge transfer that regularly transfers client knowledge to the server's global model through knowledge distillation.
However, the current methods \cite{gao2022model} for KD need each client to share a set of basic samples to make distilled soft predictions. To remedy the data leakage caused by clients sharing input samples for KD. The work \cite{zhang2023towards} proposed a federated distillation framework. 
The framework enables data-agnostic knowledge transmission between servers and clients via distributed {\em Generative Adversarial Networks} (GAN).
A PT-based method extracts sub-models from the global server model \cite{wu2023model}. 
FedRolex \cite{alam2022fedrolex} is a partial training-based approach that achieves model heterogeneity in FL, such that it enables a larger global training model compared to the local model. 
We find that current methods for solving FL heterogeneous models are suitable for HFL, meaning that each participant owns sample labels and can independently carry out local model training.
However, the passive party of VFL typically lacks labels and cannot independently train a local model.
These methods may not always apply to solving the heterogeneous model problems in VFL.

\subsection{Embedding Learning}
A typical embedding learning \cite{yang2018robust, li2021adaptive} mappings samples to a feature space and utilizes sample embeddings for classification or clustering tasks.
An embedding represents a class by computing the average of feature vectors within each class, which enables effective classification or intra-cluster compactness by maximizing the distance between embeddings \cite{michieli2021continual, wu2023federated, 10243025}.
In recent years, embeddings has been extensively explored in various domains.
In classification tasks \cite{snell2017prototypical, michieli2021continual} and clustering tasks \cite{huang2022learning}, embeddings represent a class by computing the average of feature vectors within each class, enabling effective classification or intra-cluster compactness by maximizing the distance between embeddings.
In clustering tasks, ProPos \cite{huang2022learning} aims to enhance representation uniformity and intra-cluster compactness by maximizing the distance between embeddings.
In FL embeddings are used to address data heterogeneity issues.
FedNH \cite{dai2023tackling} utilizes uniform and semantic class embeddings to tackle class imbalance and improve local models' personalization and generalization.
The work \cite{tan2022fedproto} and \cite{tan2022federated, li2021adaptive} adopt embeddings to represent classes and use average federated embedding aggregation to improve the efficiency of model training.
MPFed \cite{qiao2023framework} uses multiple embeddings to represent a class and performs model inference by testing the distance between the target embedding and multiple embeddings.
In our work, we focus on VFL in which data from all parties are aligned.
We use a single-class embedding to represent each participant's local knowledge.
By aggregating the local embeddings (e.g., embedding), we obtain knowledge from all participants that do not include model-specific details.

\section{Conclusions} \label{sec: coc}
This paper proposed a novel scheme, EASTER, to address the challenge of poor model performance due to the party-local model heterogeneity.
The key idea of EASTER was to leverage the aggregation of local embeddings rather than intermediate results to capture the local knowledge of all participants.
To ensure data privacy, we employed secure aggregation techniques to obtain global embeddings while preserving the confidentiality of the participants' original data.
The research results demonstrated that EASTER can provide simultaneous training for multiple local heterogeneous models that exhibit good performance.
In addition, our method's practical applicability and advantages contribute to the development and broader adoption of VFL in real-world scenarios.

\section*{Acknowledgments}
This work is supported by the National Key Research and Development Program of China (Grant No.s 2023YFF0905300), and the National Natural Science Foundation of China (Grant No.s U24B20146, 62372044), and Beijing Municipal Science and Technology Commission Project (Z241100009124008).
Keke Gai (gaikeke@bit.edu.cn) is a co-first author; Jing Yu (jing.yu@muc.edu.cn) is the corresponding author.

\bibliographystyle{IEEEtran}
\bibliography{ref}

\begin{thebibliography}{10}
\providecommand{\url}[1]{#1}
\csname url@samestyle\endcsname
\providecommand{\newblock}{\relax}
\providecommand{\bibinfo}[2]{#2}
\providecommand{\BIBentrySTDinterwordspacing}{\spaceskip=0pt\relax}
\providecommand{\BIBentryALTinterwordstretchfactor}{4}
\providecommand{\BIBentryALTinterwordspacing}{\spaceskip=\fontdimen2\font plus
\BIBentryALTinterwordstretchfactor\fontdimen3\font minus \fontdimen4\font\relax}
\providecommand{\BIBforeignlanguage}[2]{{%
\expandafter\ifx\csname l@#1\endcsname\relax
\typeout{** WARNING: IEEEtran.bst: No hyphenation pattern has been}%
\typeout{** loaded for the language `#1'. Using the pattern for}%
\typeout{** the default language instead.}%
\else
\language=\csname l@#1\endcsname
\fi
#2}}
\providecommand{\BIBdecl}{\relax}
\BIBdecl

\bibitem{pan2023fedmdfg}
Z.~Pan, S.~Wang, C.~Li, H.~Wang, X.~Tang, and J.~Zhao, ``{FedMDFG}: Federated learning with multi-gradient descent and fair guidance,'' in \emph{Proceedings of the AAAI Conference on Artificial Intelligence}, 2023, pp. 9364--9371.

\bibitem{so2023securing}
J.~So, R.~E. Ali, B.~G{\"u}ler, and et~al, ``Securing secure aggregation: Mitigating multi-round privacy leakage in federated learning,'' in \emph{Proceedings of the AAAI Conference on Artificial Intelligence}, 2023, pp. 9864--9873.

\bibitem{gong2023multi}
M.~Gong, Y.~Zhang, Y.~Gao, A.~K. Qin, Y.~Wu, S.~Wang, and Y.~Zhang, ``A multi-modal vertical federated learning framework based on homomorphic encryption,'' \emph{IEEE Transactions on Information Forensics and Security}, vol.~19, pp. 1826--1839, 2024.

\bibitem{aledhari2020federated}
M.~Aledhari, R.~Razzak, R.~M. Parizi, and F.~Saeed, ``Federated learning: A survey on enabling technologies, protocols, and applications,'' \emph{IEEE Access}, vol.~8, pp. 140\,699--140\,725, 2020.

\bibitem{li2021survey}
Q.~Li, Z.~Wen, Z.~Wu, and et~al, ``A survey on federated learning systems: Vision, hype and reality for data privacy and protection,'' \emph{IEEE Transactions on Knowledge and Data Engineering}, vol.~35, no.~4, pp. 3347--3366, 2023.

\bibitem{xia2022cascade}
W.~Xia, Y.~Li, L.~Zhang, Z.~Wu, and X.~Yuan, ``Cascade vertical federated learning,'' in \emph{2022 IEEE International Conference on Multimedia and Expo (ICME)}, 2022, pp. 1--6.

\bibitem{wu2023falcon}
Y.~Wu, N.~Xing, G.~Chen, and et~al, ``Falcon: A privacy-preserving and interpretable vertical federated learning system,'' \emph{Proceedings of the VLDB Endowment}, vol.~16, no.~10, pp. 2471--2484, 2023.

\bibitem{jin2021cafe}
X.~Jin, P.-Y. Chen, C.-Y. Hsu, and et~al, ``Catastrophic data leakage in vertical federated learning,'' in \emph{Advances in Neural Information Processing Systems}, 2021, pp. 994--1006.

\bibitem{wei2023fedads}
P.~Wei, H.~Dou, S.~Liu, R.~Tang, L.~Liu, L.~Wang, and B.~Zheng, ``{FedAds}: {A} benchmark for privacy-preserving {CVR} estimation with vertical federated learning,'' in \emph{Proceedings of the 46th International ACM SIGIR Conference on Research and Development in Information Retrieval}, 2023, pp. 3037--3046.

\bibitem{castiglia2022compressed}
T.~J. Castiglia, A.~Das, S.~Wang, and S.~Patterson, ``Compressed-{VFL}: Communication-efficient learning with vertically partitioned data,'' in \emph{International Conference on Machine Learning}, 2022, pp. 2738--2766.

\bibitem{liu2022vertical}
R.~Xu, N.~Baracaldo, Y.~Zhou, and et~al, ``Fedv: Privacy-preserving federated learning over vertically partitioned data,'' in \emph{Proceedings of the 14th ACM workshop on artificial intelligence and security}, 2021, pp. 181--192.

\bibitem{liu2024vertical}
Y.~Liu, Y.~Kang, T.~Zou, Y.~Pu, Y.~He, X.~Ye, Y.~Ouyang, Y.-Q. Zhang, and Q.~Yang, ``Vertical federated learning: Concepts, advances, and challenges,'' \emph{IEEE Transactions on Knowledge and Data Engineering}, vol.~36, no.~7, pp. 3615--3634, 2024.

\bibitem{ye2023heterogeneous}
M.~Ye, X.~Fang, B.~Du, P.~C. Yuen, and D.~Tao, ``Heterogeneous federated learning: State-of-the-art and research challenges,'' \emph{ACM Computing Surveys}, vol.~56, no.~3, pp. 1--44, 2023.

\bibitem{tan2022fedproto}
Y.~Tan, G.~Long, L.~Liu, T.~Zhou, Q.~Lu, J.~Jiang, and C.~Zhang, ``Fedproto: Federated prototype learning across heterogeneous clients,'' in \emph{Proceedings of the AAAI Conference on Artificial Intelligence}, 2022, pp. 8432--8440.

\bibitem{fang2022robust}
X.~Fang and M.~Ye, ``Robust federated learning with noisy and heterogeneous clients,'' in \emph{Proceedings of the IEEE/CVF Conference on Computer Vision and Pattern Recognition}, 2022, pp. 10\,072--10\,081.

\bibitem{wang2023flexifed}
K.~Wang, Q.~He, F.~Chen, and et~al, ``{FlexiFed}: Personalized federated learning for edge clients with heterogeneous model architectures,'' in \emph{Proceedings of the ACM Web Conference 2023}, 2023, pp. 2979--2990.

\bibitem{qi2024model}
P.~Qi, D.~Chiaro, A.~Guzzo, M.~Ianni, G.~Fortino, and F.~Piccialli, ``Model aggregation techniques in federated learning: A comprehensive survey,'' \emph{Future Generation Computer Systems}, vol. 150, no.~99, pp. 272--293, 2024.

\bibitem{qiao2023prototype}
Y.~Qiao, S.-B. Park, S.~M. Kang, and et~al, ``Prototype helps federated learning: Towards faster convergence,'' \emph{arXiv}, vol.~PP, no.~99, p.~1, 2023.

\bibitem{ma2023flamingo}
Y.~Ma, J.~Woods, S.~Angel, and et~al, ``Flamingo: Multi-round single-server secure aggregation with applications to private federated learning,'' in \emph{2023 IEEE Symposium on Security and Privacy (SP)}, 2023, pp. 477--496.

\bibitem{ye2024vertical}
M.~Ye, W.~Shen, B.~Du, E.~Snezhko, V.~Kovalev, and P.~C. Yuen, ``Vertical federated learning for effectiveness, security, applicability: A survey,'' \emph{ACM Computing Surveys}, vol.~21, no.~9, pp. 1 -- 32, 2024.

\bibitem{li2024approaching}
L.~Li, K.~Hu, X.~Zhu, S.~Jiang, L.~Weng, and M.~Xia, ``Approaching expressive and secure vertical federated learning with embedding alignment in intelligent iot systems,'' \emph{IEEE Internet of Things Journal}, vol.~11, no.~23, pp. 38\,657 -- 38\,672, 2024.

\bibitem{zhang2021secure}
Q.~Zhang, B.~Gu, C.~Deng, and H.~Huang, ``Secure bilevel asynchronous vertical federated learning with backward updating,'' in \emph{Proceedings of the AAAI Conference on Artificial Intelligence}, 2021, pp. 10\,896--10\,904.

\bibitem{citation-key}
S.~Li, D.~Yao, and J.~Liu, ``{FedVS}: Straggler-resilient and privacy-preserving vertical federated learning for split models,'' in \emph{International Conference on Machine Learning}, 2023, pp. 20\,296--20\,311.

\bibitem{tan2022federated}
Y.~Tan, G.~Long, J.~Ma, L.~LIU, T.~Zhou, and J.~Jiang, ``{Federated Learning from Pre-Trained Models: A Contrastive Learning Approach},'' in \emph{Advances in Neural Information Processing Systems}.\hskip 1em plus 0.5em minus 0.4em\relax Curran Associates, Inc., 2022, pp. 19\,332--19\,344.

\bibitem{gao2021convergence}
H.~Gao, A.~Xu, and H.~Huang, ``On the convergence of communication-efficient local {SGD} for federated learning,'' in \emph{Proceedings of the AAAI Conference on Artificial Intelligence}, 2021, pp. 7510--7518.

\bibitem{zheng2022aggregation}
Y.~Zheng, S.~Lai, Y.~Liu, and et~al, ``Aggregation service for federated learning: An efficient, secure, and more resilient realization,'' \emph{IEEE Transactions on Dependable and Secure Computing}, vol.~20, no.~2, pp. 988--1001, 2023.

\bibitem{romanini2021pyvertical}
D.~Romanini, A.~J. Hall, P.~Papadopoulos, and et~al, ``Pyvertical: A vertical federated learning framework for multi-headed {splitNN},'' \emph{arXiv}, vol.~PP, no.~99, p.~1, 2021.

\bibitem{zhang2022adaptive}
J.~Zhang, S.~Guo, Z.~Qu, D.~Zeng, H.~Wang, Q.~Liu, and A.~Y. Zomaya, ``Adaptive vertical federated learning on unbalanced features,'' \emph{IEEE Transactions on Parallel and Distributed Systems}, vol.~33, no.~12, pp. 4006--4018, 2022.

\bibitem{zouvflair}
T.~Zou, G.~Zixuan, Y.~He, H.~Takahashi, Y.~Liu, and Y.-Q. Zhang, ``Vflair: A research library and benchmark for vertical federated learning,'' in \emph{The Twelfth International Conference on Learning Representations}, Vienna Austria, 2024, pp. 1--39.

\bibitem{lecun1998gradient}
Y.~LeCun, L.~Bottou, Y.~Bengio, and P.~Haffner, ``Gradient-based learning applied to document recognition,'' \emph{Proceedings of the IEEE}, vol.~86, no.~11, pp. 2278--2324, 1998.

\bibitem{xiao2017fashion}
X.~Han, R.~Kashif, and V.~Roland, ``{Fashion-MNIST}: a novel image dataset for benchmarking machine learning algorithms,'' \emph{CoRR}, vol.~PP, no.~99, p.~1, 2017.

\bibitem{krizhevsky2009learning}
A.~Krizhevsky, \emph{Learning multiple layers of features from tiny images}.\hskip 1em plus 0.5em minus 0.4em\relax ON, Canada: Toronto, 2009.

\bibitem{NEURIPS2021_2f2b2656}
M.~Luo, F.~Chen, D.~Hu, Y.~Zhang, J.~Liang, and J.~Feng, ``No fear of heterogeneity: Classifier calibration for federated learning with non-iid data,'' in \emph{Advances in Neural Information Processing Systems}, M.~Ranzato, A.~Beygelzimer, Y.~Dauphin, P.~Liang, and J.~W. Vaughan, Eds., vol.~34, 2021, pp. 5972--5984.

\bibitem{deng2009imagenet}
J.~Deng, W.~Dong, R.~Socher, L.-J. Li, K.~Li, and L.~Fei-Fei, ``Imagenet: A large-scale hierarchical image database,'' in \emph{2009 IEEE Conference on Computer Vision and Pattern Recognition}, 2009, pp. 248--255.

\bibitem{10380676}
F.~Fu, X.~Wang, J.~Jiang, H.~Xue, and B.~Cui, ``Projpert: Projection-based perturbation for label protection in split learning based vertical federated learning,'' \emph{IEEE Transactions on Knowledge and Data Engineering}, vol.~36, no.~7, pp. 3417--3428, 2024.

\bibitem{taud2018multilayer}
P.~Moallem and S.~A. Ayoughi, ``Removing potential flat spots on error surface of multilayer perceptron {(MLP)} neural networks,'' \emph{International Journal of Computer Mathematics}, vol.~88, no.~1, pp. 21--36, 2011.

\bibitem{albawi2017understanding}
S.~Albawi, T.~A. Mohammed, and S.~Al-Zawi, ``Understanding of a convolutional neural network,'' in \emph{2017 International Conference on Engineering and Technology}, 2017, pp. 1--6.

\bibitem{lecun2015lenet}
Y.~LeCun \emph{et~al.}, ``{LeNet-5}, convolutional neural networks,'' \emph{URL: http://yann.lecun.com/exdb/lenet}, vol.~20, no.~5, p.~14, 2015.

\bibitem{subaar2024investigating}
C.~Subaar, F.~T. Addai, E.~C.~K. Addison, O.~Christos, J.~Adom, M.~Owusu-Mensah, N.~Appiah-Agyei, and S.~Abbey, ``Investigating the detection of breast cancer with deep transfer learning using resnet18 and resnet34,'' \emph{Biomedical Physics \& Engineering Express}, vol.~10, no.~3, p. 035029, 2024.

\bibitem{guo2017deepfm}
H.~Guo, R.~Tang, Y.~Ye, Z.~Li, and X.~He, ``Deepfm: a factorization-machine based neural network for ctr prediction,'' \emph{arXiv preprint arXiv:1703.04247}, 2017.

\bibitem{cheng2016wide}
H.-T. Cheng, L.~Koc, J.~Harmsen, T.~Shaked, T.~Chandra, H.~Aradhye, G.~Anderson, G.~Corrado, W.~Chai, M.~Ispir \emph{et~al.}, ``Wide \& deep learning for recommender systems,'' in \emph{Proceedings of the 1st workshop on deep learning for recommender systems}, Boston, MA, USA, 2016, pp. 7--10.

\bibitem{paszke2019pytorch}
P.~Adam, G.~Sam, M.~Francisco, L.~Adam, B.~James, C.~Gregory, K.~Trevor \emph{et~al.}, ``{PyTorch}: An imperative style, high-performance deep learning library,'' in \emph{Advances in Neural Information Processing Systems}, 2019, pp. 8024--8035.

\bibitem{huang2023incentive}
J.~Huang, B.~Ma, M.~Wang, X.~Zhou, L.~Yao, S.~Wang, L.~Qi, and Y.~Chen, ``Incentive mechanism design of federated learning for recommendation systems in {MEC},'' \emph{{IEEE} Trans. Consumer Electron.}, vol.~70, no.~1, pp. 2596--2607, 2024.

\bibitem{chatterjee2023federated}
P.~Chatterjee, D.~Das, and D.~B. Rawat, ``Federated learning empowered recommendation model for financial consumer services,'' \emph{{IEEE} Trans. Consumer Electron.}, vol.~70, no.~1, pp. 2508 -- 2516, 2023.

\bibitem{nguyen2022federated}
D.~C. Nguyen, Q.-V. Pham, P.~N. Pathirana, M.~Ding, A.~Seneviratne, Z.~Lin, O.~Dobre, and W.-J. Hwang, ``Federated learning for smart healthcare: A survey,'' \emph{{ACM} Comput. Surv.}, vol.~55, no.~3, pp. 1--37, 2022.

\bibitem{wen2023task}
D.~Wen, P.~Liu, G.~Zhu, Y.~Shi, J.~Xu, Y.~C. Eldar, and S.~Cui, ``Task-oriented sensing, computation, and communication integration for multi-device edge ai,'' \emph{IEEE Transactions on Wireless Communications}, vol.~23, no.~3, pp. 2486--2502, 2023.

\bibitem{hu2021mhat}
L.~Hu, H.~Yan, L.~Li, Z.~Pan, X.~Liu, and Z.~Zhang, ``{MHAT}: An efficient model-heterogenous aggregation training scheme for federated learning,'' \emph{Information Sciences}, vol. 560, no.~99, pp. 493--503, 2021.

\bibitem{huang2022learn}
W.~Huang, M.~Ye, and B.~Du, ``Learn from others and be yourself in heterogeneous federated learning,'' in \emph{Proceedings of the IEEE/CVF Conference on Computer Vision and Pattern Recognition}, 2022, pp. 10\,143--10\,153.

\bibitem{zhu2021data}
Z.~Zhu, J.~Hong, and J.~Zhou, ``Data-free knowledge distillation for heterogeneous federated learning,'' in \emph{International conference on machine learning}, 2021, pp. 12\,878--12\,889.

\bibitem{cho2023communication}
Y.~J. Cho, J.~Wang, T.~Chirvolu, and G.~Joshi, ``Communication-efficient and model-heterogeneous personalized federated learning via clustered knowledge transfer,'' \emph{IEEE Journal of Selected Topics in Signal Processing}, vol.~17, no.~1, pp. 234--247, 2023.

\bibitem{zhang2023towards}
J.~Zhang, S.~Guo, J.~Guo, D.~Zeng, J.~Zhou, and A.~Zomaya, ``Towards data-independent knowledge transfer in model-heterogeneous federated learning,'' \emph{IEEE Transactions on Computers}, vol.~72, no.~10, pp. 2888--2901, 2023.

\bibitem{he2020group}
C.~He, M.~Annavaram, and S.~Avestimehr, ``Group knowledge transfer: Federated learning of large cnns at the edge,'' in \emph{Advances in Neural Information Processing Systems}, 2020, pp. 14\,068--14\,080.

\bibitem{alam2022fedrolex}
S.~Alam, L.~Liu, M.~Yan, and M.~Zhang, ``{FedRolex}: Model-heterogeneous federated learning with rolling sub-model extraction,'' in \emph{Advances in Neural Information Processing Systems}, 2022, pp. 29\,677--29\,690.

\bibitem{gao2022model}
T.~Gao, X.~Jin, and Y.~Lai, ``Model heterogeneous federated learning for intrusion detection based on knowledge distillation,'' in \emph{Proceedings of the 2022 12th International Conference on Communication and Network Security}.\hskip 1em plus 0.5em minus 0.4em\relax Beijing, China: {ACM}, 2022, pp. 30--35.

\bibitem{wu2023model}
H.~Wu, P.~Wang, and A.~C. Narayan, ``Model-heterogeneous federated learning with partial model training,'' in \emph{2023 IEEE/CIC International Conference on Communications in China (ICCC)}, 2023, pp. 1--6.

\bibitem{yang2018robust}
H.~Yang, X.~Zhang, F.~Yin, and C.~Liu, ``Robust classification with convolutional prototype learning,'' in \emph{Proceedings of the IEEE/CVF Conference on Computer Vision and Pattern Recognition}, 2018, pp. 3474--3482.

\bibitem{li2021adaptive}
G.~Li, V.~Jampani, L.~Sevilla-Lara, D.~Sun, J.~Kim, and J.~Kim, ``Adaptive prototype learning and allocation for few-shot segmentation,'' in \emph{Proceedings of the IEEE/CVF Conference on Computer Vision and Pattern Recognition}, 2021, pp. 8334--8343.

\bibitem{michieli2021continual}
U.~Michieli and P.~Zanuttigh, ``Continual semantic segmentation via repulsion-attraction of sparse and disentangled latent representations,'' in \emph{Proceedings of the IEEE/CVF conference on computer vision and pattern recognition}, 2021, pp. 1114--1124.

\bibitem{wu2023federated}
H.~Wu, B.~Zhang, C.~Chen, and J.~Qin, ``Federated semi-supervised medical image segmentation via prototype-based pseudo-labeling and contrastive learning,'' \emph{IEEE Transactions on Medical Imaging}, vol.~PP, no.~99, p.~1, 2023.

\bibitem{10243025}
A.~Wang, L.~Yang, H.~Wu, and Y.~Iwahori, ``Heterogeneous defect prediction based on federated prototype learning,'' \emph{IEEE Access}, vol.~11, pp. 98\,618--98\,632, 2023.

\bibitem{snell2017prototypical}
J.~Snell, K.~Swersky, and R.~Zemel, ``Prototypical networks for few-shot learning,'' in \emph{Advances in Neural Information Processing Systems}, 2017, pp. 1--11.

\bibitem{huang2022learning}
Z.~Huang, J.~Chen, J.~Zhang, and H.~Shan, ``Learning representation for clustering via prototype scattering and positive sampling,'' \emph{{IEEE} Transactions on Pattern Analysis and Machine Intelligence}, vol.~45, no.~6, pp. 7509--7524, 2022.

\bibitem{dai2023tackling}
Y.~Dai, Z.~Chen, J.~Li, S.~Heinecke, L.~Sun, and R.~Xu, ``Tackling data heterogeneity in federated learning with class prototypes,'' in \emph{Proceedings of the AAAI Conference on Artificial Intelligence}, 2023, pp. 7314--7322.

\bibitem{qiao2023framework}
Y.~Qiao, M.~S. Munir, A.~Adhikary, A.~D. Raha, S.~H. Hong, and C.~S. Hong, ``A framework for multi-prototype based federated learning: Towards the edge intelligence,'' in \emph{International Conference on Information Processing in Sensor Networks}, 2023, pp. 134--139.

\end{thebibliography}

\begin{IEEEbiography}[{\includegraphics[width=1in,height=1.25in, clip,keepaspectratio]{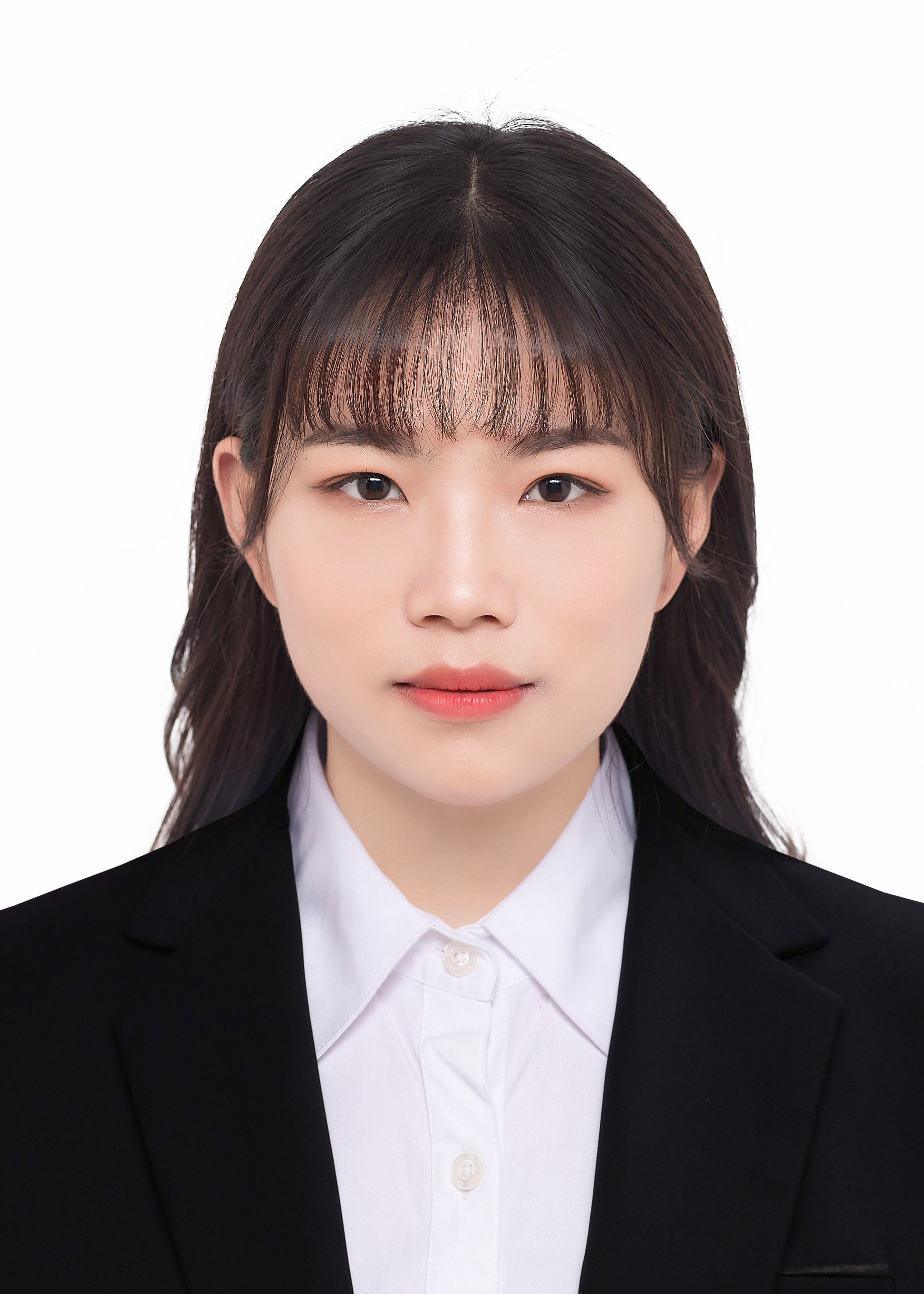}}]{Shuo Wang} (Student Member, IEEE)
    is currently pursuing a Ph.D. degree in Electronic Information at the School of Cyberspace Science and Technology, Beijing Institute of Technology, Beijing, China.
     Her current research interests include federated learning, privacy-preserving computation, and artificial intelligence security.
\end{IEEEbiography}

\begin{IEEEbiography}[{\includegraphics[width=1in,height=1.25in, clip,keepaspectratio]{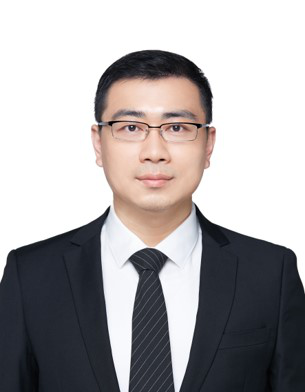}}]{Keke Gai} (Senior Member, IEEE) received the Ph.D. degree in computer science from Pace University, New York, NY, USA. 
% He is currently a Professor at the School of Artificial Intelligence, and at the School of Cyberspace Science and Technology, Beijing Institute of Technology, Beijing , China.
He is currently a Professor at both the School of Artificial Intelligence and the School of Cyberspace Science and Technology, Beijing Institute of Technology, Beijing, China.
He has published 5 books, more than 180 peer-reviewed papers.
He serves as an Editor-in-Chief of the journal Blockchains and Associate Editors for a few journals, including IEEE Transactions on Dependable and Secure Computing, Journal of Parallel and Distributed Computing, etc.
He also serves as a co-chair of IEEE Technology and Engineering Management Society's Technical Committee on Blockchain and Distributed Ledger Technologies, a Standing Committee Member at China Computer Federation-Blockchain Committee.
His research interests include cybersecurity, blockchain, privacy-preserving computation, and decentralized identity.
\end{IEEEbiography}

% \IEEEcompsocthanksitem K. Gai is with the School of Artificial Intelligence, Beijing Institute of Technology, Beijing, China, and is also with the School of Cyberspace Science and Technology, Beijing Institute of Technology, Beijing 100081, China. (e-mail: gaikeke@bit.edu.cn).
% \IEEEcompsocthanksitem J. Yu is with the Key Laboratory of Ethnic Language Intelligent Analysis and Security Governance of MOE, Minzu University of China, and is also with the School of Information Engineering, Minzu University of China. (Email: jing.yu@muc.edu.cn).

\begin{IEEEbiography}[{\includegraphics[width=1in,height=1.25in,clip,keepaspectratio]{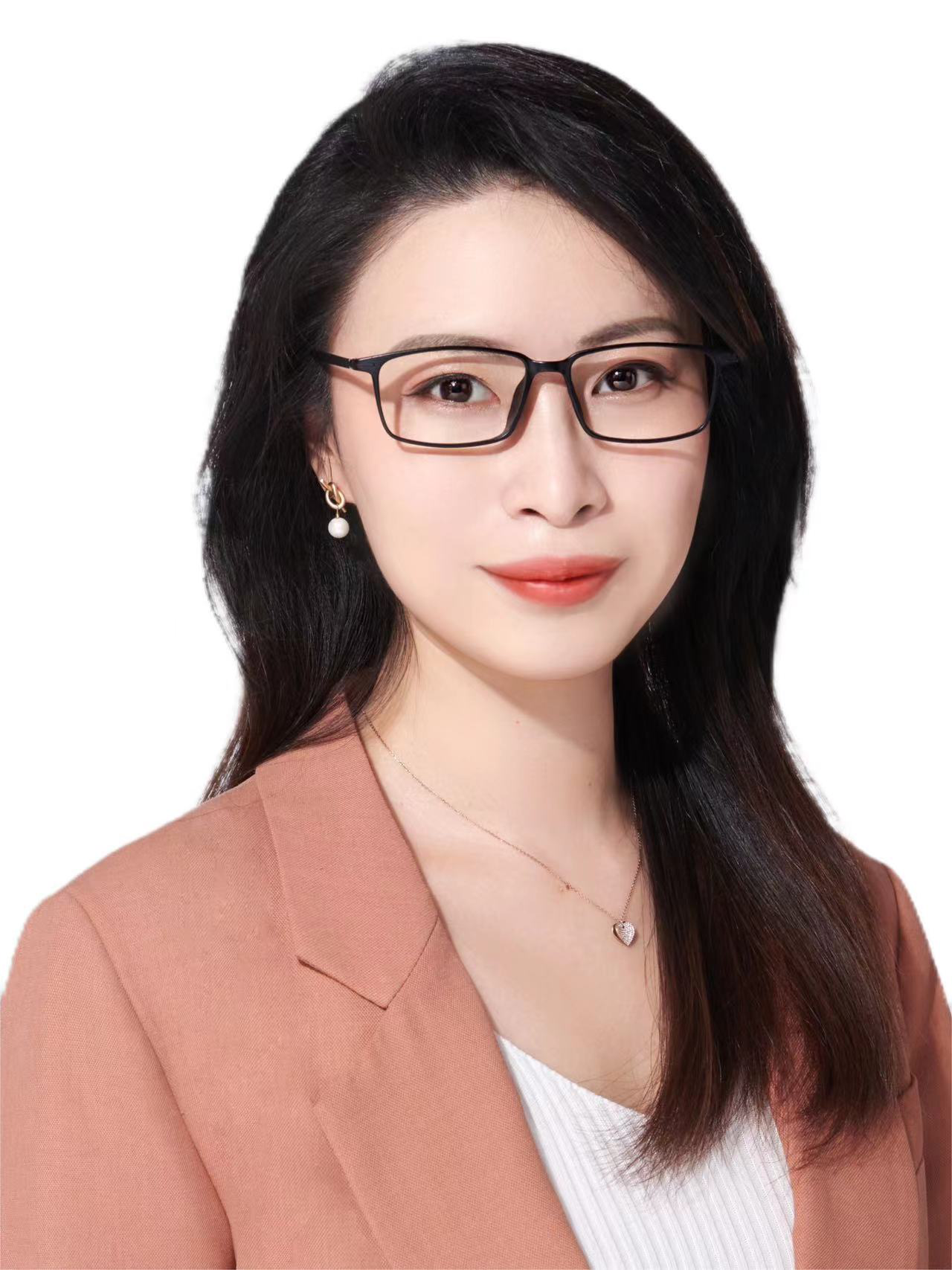}}]{Jing Yu} (Member, IEEE) is currently an associate professor at both the Key Laboratory of Ethnic Language Intelligent Analysis and Security Governance of MOE and the School of Information Engineering, Minzu University of China, Beijing, China. 
Jing Yu received her B.S. degree in Automation Science from Minzu University, China, in 2011, and got her  M.S. degree in Pattern Recognition from Beihang University, China in 2014. She received her Ph.D.  degree in the University of Chinese Academy of Sciences, China, in 2019. Her research interests mainly focus on cross-modal understanding and artificial intelligence security.
\end{IEEEbiography}

\begin{IEEEbiography}[{\includegraphics[width=1in,height=1.25in,clip,keepaspectratio]{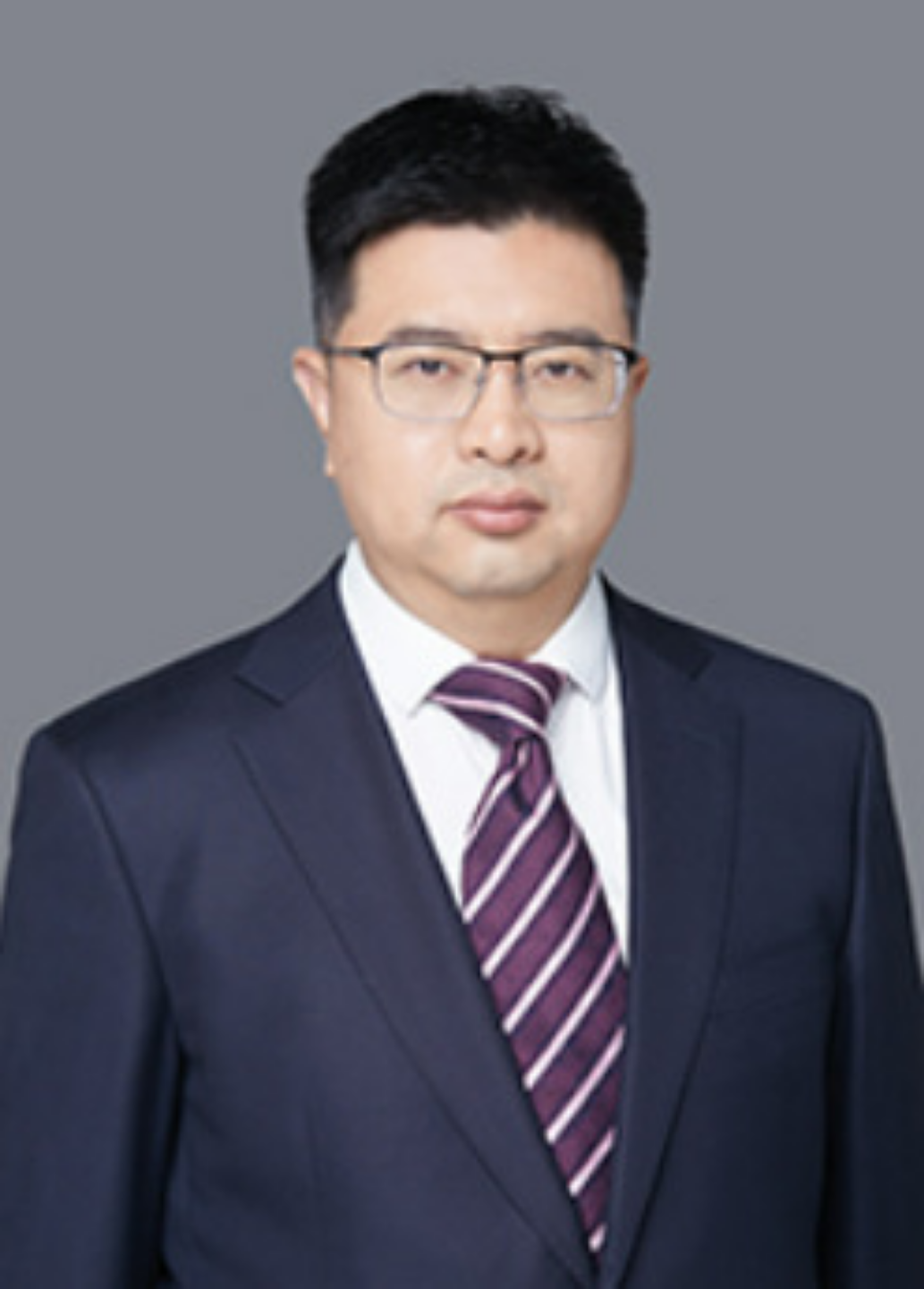}}]{Liehuang Zhu}
	 (Senior Member, IEEE) is currently a Full Professor with the School of Cyberspace Science and Technology, Beijing Institute of Technology.
  He is selected into the Program for New Century Excellent Talents in University from Ministry of Education, China. He has published over 60 SCI-indexed research articles in these areas, and a book published by Springer.  
  He serves on the editorial boards of three international journals, including IEEE Internet of Things Journal, IEEE Transactions on Vehicular Technology, and IEEE Network magazine. 
  He won the Best Paper Award at IEEE/ACM IWQoS 2017 and IEEE TrustCom 2018.
  His research interests include the Internet of Things, cloud computing security, internet, and mobile security.
\end{IEEEbiography}

\begin{IEEEbiography}[{\includegraphics[width=1in,height=1.25in,clip,keepaspectratio]{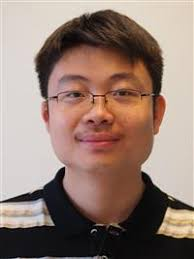}}]{Weizhi Meng}
   (Senior Member, IEEE) is a Full Professor in the School of Computing and Communications, Lancaster University, United Kingdom, and an adjunct faculty in the Department of Applied Mathematics and Computer Science, Technical University of Denmark, Denmark. He obtained his Ph.D. degree in Computer Science from the City University of Hong Kong. He was a recipient of the Hong Kong Institution of Engineers (HKIE) Outstanding Paper Award for Young Engineers/Researchers in both 2014 and 2017. He also received the IEEE ComSoc Best Young Researcher Award for Europe, Middle East, \& Africa Region (EMEA) in 2020 and the IEEE ComSoc Communications \& Information Security (CISTC) Early Career Award in 2023. His primary research interests are blockchain technology, cyber security and artificial intelligence in security including intrusion detection, blockchain applications, smartphone security, biometric authentication, and IoT security. He is senior member of IEEE.
\end{IEEEbiography}

\begin{IEEEbiography}[{\includegraphics[width=1in,height=1.25in,clip,keepaspectratio]{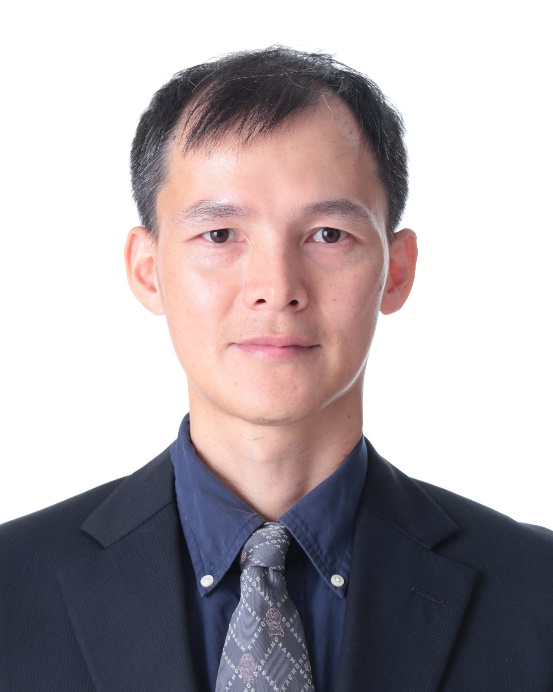}}]{Bin Xiao} (Fellow, IEEE)
is a professor at the Department of Computing, the Hong Kong Polytechnic University, Hong Kong. Prof. Xiao received B.Sc and M.Sc degrees in Electronics Engineering from Fudan University, China, and a Ph.D. degree in computer science from the University of Texas at Dallas, USA. His research interests include AI security and privacy, data privacy, Web3, and blockchain systems. He published more than 200 technical papers in international top journals and conferences. He is currently an Associate Editor of the IEEE Transactions on Cloud Computing. He has been the associate editor of the IEEE Internet of Things Journal, IEEE Transactions on Network Science and Engineering, and Elsevier Journal of Parallel and Distributed Computing. He has been the chair of the IEEE ComSoc CISTC committee from 2024 to 2025 and the IEEE ComSoc Distinguished Lecturer from 2023 to 2024. He has been the program co-chair of IEEE CNS2025, track co-chair of IEEE ICDCS2022, the symposium track co-chair of IEEE Globecom 2024, ICC2020, ICC 2018, and Globecom 2017, and the general chair of IEEE SECON 2018. 
\end{IEEEbiography}

\end{document}